\documentclass{article}

\usepackage{amsmath}
\usepackage{amssymb}
\usepackage[round]{natbib}
\usepackage{color}
\usepackage{graphicx}
\usepackage{caption}
\usepackage{subcaption}
\usepackage{float}
\usepackage{lastpage}
\usepackage{chngcntr}
\counterwithin{figure}{section}
\usepackage{array}
\usepackage{titling}

\title{Infinite Structured Sparse Factor Analysis}
\author{Matthew C. Pearce \\ MRC Biostatistics Unit \and Cam-CAN \and Simon R. White \\ MRC Biostatistics Unit}

\begin{document}

\maketitle

\begin{abstract}
Matrix factorisation methods decompose multivariate observations as linear combinations of latent feature vectors. The Indian Buffet Process (IBP) provides a way to model the number of latent features required for a good approximation in terms of regularised reconstruction error. Previous work has focussed on latent feature vectors with independent entries. We extend the model to include nondiagonal latent covariance structures representing characteristics such as smoothness. This is done by . Using simulations we demonstrate that under appropriate conditions a smoothness prior helps to recover the true latent features, while denoising more accurately. We demonstrate our method on a real neuroimaging dataset, where computational tractability is a sufficient challenge that the efficient strategy presented here is essential.
\end{abstract}

\section{Introduction}
\label{sec:istica}

Matrix factorisation methods decompose multivariate observations into linear combinations of latent features. Consider a $(T \times V)$ data matrix $\bold Y$ comprising $T$ observations of $V$ dimensional vectors, which is decomposed into a linear combinations of $K$ features as:
\begin{flalign}
  & \bold Y_{(T \times V)} = \bold W_{(T \times K)} \bold S_{(K \times V)} + \bold E_{(T \times V)}
\label{eq:generaldecomp}\end{flalign}

Dimension reduction can then be effected by approximating $\bold Y$ with $\bold W \bold S$, thus using a smaller number of variables. Here $\bold E$ is the difference between the low rank approximation and the data, whose sum of squares is the reconstruction error. We will refer to the entries of matrices playing the role of $\bold W$ as weights and rows of $\bold S$ as features. A family of methods can be placed within this algebraic structure, including Principal Components Analysis (PCA), Factor Analysis (FA), Independent Components Analysis (ICA) and Non-negative Matrix Factorisation (NMF). 


When applying matrix decomposition we generally face uncertainty about the number of latent features, $K$, required to explain the data. We start from the assumption that the reader wishes to learn the latent features, their weights and their number from the data in a single, coherent model for approximating the data. Due to the wide family of methods which conform to the algebraic structure in (\ref{eq:generaldecomp}) methods which improve our accounting for dimensional uncertainty have a broad scope for application. At the same time we seek decompositions with desirable characteristics, such as sparse activation of features, or constraints on the covariance structure of latent features. 


A previous approach to accounting for uncertainty in $K$ was proposed by \citet{viroli2009reversibleFA}. In that scheme the weights assigned to feature vectors were draws from Gaussian mixture models, while the dimension $K$ was sampled through reversible jump MCMC (RJMCMC). Such a model implies that the representation of observations with respect to the latent features would be dense. 

We will primarily be concerned with observations that are natural images. Work on natural images (\citet{hyvarinen2009natural}), together with successes in signal processing (\citet{elad2010role}) and representation learning (\citet{bengio2013representation}) suggests that modelling representations of images as sparse with respect to a basis is a powerful and parsimonious strategy. In the formalism of (\ref{eq:generaldecomp}) this means $\bold W$ being sparse when $\bold S$ was chosen appropriately. Hence the approach in \citet{viroli2009reversibleFA} would not be entirely satisfying as a model for natural images.

An alternate strategy to account for uncertainty in $K$ was proposed by \citet{knowles2007infinite} through use of the Indian Buffet Process (IBP).  A draw, $\bold Z$, from the $ IBP(\alpha, \beta)$ distribution is a binary matrix and can be used in an elementwise product (denoted $\circ$) with a matrix of scaling coefficents $\bold A$ to produce a sparse, random, weighting matrix, $\bold W = \bold A \circ \bold Z$. Here $\bold A$ and $\bold Z$ are theoretically $(T \times \infty)$ but, for finite $T$, $\bold Z$ is guaranteed to have a finite number of nonzero columns, making computational use tractable. The decomposition in \ref{eq:generaldecomp} becomes:
\begin{flalign} \label{eqn:decomp}
  & \bold Y = (\bold A \circ \bold Z) \bold S + \bold E 
\end{flalign}
\citet{knowles2007infinite} provides models for infinite Independent Components Analysis (iICA) and infinite sparse Factor Analysis (isFA) by changing the elementwise distributional assumptions on entries of $\bold A$ as following a Laplace or Gaussian distribution respectively. Note that when dealing with the IBP that the effective rank of the decomposition, $K$, is merely a statistic arising by counting the nonzero columns in $\bold Z$. This means $K$ is not a parameter to be sampled separately, and so specialist RJMCMC techniques are not required.

A limitation of the existing work is the use of IID univariate distributions to model latent features which we reasonably expect to exhibit strong dependency structures. That is, existing models provide a route to sparsity with respect to a basis, but do not exploit prior beliefs about the structure of that basis. For instance a collection of time series that were sparse in the Fourier basis would have features in the rows of $\bold S$ that were smooth. We attend to this problem in Section \ref{sec:textures} by considering efficient models for latent covariance structure via parameterised eigendecompositions.  We further show that these models can be designed with interpretable representations as Gaussian Markov Random Fields (GMRFs) (\citet{rue2005gmrfs}). 

In Section \ref{sec:model} we present the contribution of this paper, an extended infinite Independent Structured Sparse Factor Analysis model. Section \ref{sec:sampler} provides a Gibbs sampling algorithm for our model. Our method's utility is demonstrated in Section \ref{sec:simulation} where it is found to outperform PCA and prior Bayesian nonparametric work on simulated data. We illustrate the scalability of the algorithm with a neuroimaging application in Section \ref{sec:neuroexample}. Finally, Section \ref{sec:discussion} is reserved for discussion.

\section{Structured Features}
\label{sec:textures}

Matrix factorisation models typically do not capture correlation structure within latent vectors $\bold s$, setting aside vector normalisation. For example in classical PCA, see \cite{mardia1979multivariate}, we do not expect $\bold s_i$ to have a similar value to $\bold s_{i+1}$ . This is replicated in the Bayesian PCA of \citet{tipping1999probabilistic}.  Similarly treatments such as the sparse PCA of \citet{zou2006sparse} modify the distributions of the latent variables to induce sparsity in one of the factors but without assuming that the nonzero coefficients follow a pattern (e.g. nonzeros come in blocks). Popular matrix factorisation techniques in recommender systems \citet{koren2009matrix} are similar. However for certain classes of data, e.g. locations in space, points in time, groups of people, we may expect structure within our latent features, perhaps in the form of smooth landscapes, cyclic time-courses, or group-wise random effects. 

Modelling correlation structure for latent vectors could be prohibitive for $V$ dimensional data as it implies handling at least one $V \times V$ nondiagonal covariance matrix. For instance, in fMRI neuroimaging where $V$ can be on the order of $10^5$ or $10^6$, hence related covariance matrices cannot be naively handled, even on high performance computers. For instance evaluating the density of a multivariate normal requires the determinant of the covariance matrix which in the general case is $\mathcal O (V^3)$. 

Now, suppose we knew the eigendecomposition of a precision matrix $\bold Q = \bold U \bold D \bold U^T$. Then we could evaluate the determinant of the covariance matrix in $\mathcal O(V)$ time. However, we would still be in numerical difficulty as $\bold U$ may not fit into the memory of a computer, and if it did required matrix-vector operations would still be $\mathcal O(V^2)$.

Imagine that we knew of fast linear operators which encoded the actions of $\bold U$ and $\bold U^T$, i.e. $\text{op}(\bold U) \bold v$ could be obtained in $\mathcal O(t(V))$ time for some $t(V) \ll V^2$. Then we could both evaluate the density and sample from the distribution in $\mathcal O(t(V))$ time.

We show that this is possible for several interesting classes of precision matrices. For certain important precision matrices the results are interpretable within the Gaussian Markov Random Field framework of \citet{rue2005gmrfs}. We also provide rules for their combination into richer models.

\subsection{Parameterisation}

The work presented in this paper models latent features as drawn from $\bold s \sim \mathcal N_V( \bold 0,\bold Q(\theta)^{-1})$. We consider parameterised precision matrices $\bold Q(\theta)$ of the form:
\begin{flalign*}
  & \bold Q(\theta) = \bold U \text{diag}(\bold h(\theta)) \bold U^T = \bold U \bold D(\theta) \bold U^T
\end{flalign*}
where $\theta_i > 0$, $\bold h_i(\theta)>0$ (so that $\bold Q(\theta) > 0$) and is twice continuously differentiable. It is assumed that $\bold U$ is a known orthonormal matrix and for practical purposes can be represented as a fast linear operator. We will sometimes suppress the vector parameter $\theta$ for notational convenience.

In this case we can evaluate the pdf as:
\begin{flalign*}
  & \ln f(\bold s | \theta) = c + \frac{1}{2}\sum_{v=1}^V \ln(\bold h_i(\theta)) - \frac{1}{2}\sum_{i=1}^V \bold h_i(\theta)(\bold U^T \bold s)_i^2
\end{flalign*}

We also note that this formulation is straightforwardly differentiable with respect to the parameter, enabling various optimisation and sampling procedures for its posterior distribution:
\begin{flalign*}
  & \nabla_{\theta} \ln f(\bold s | \theta)= \frac{1}{2}\sum_{i=1}^V \frac{ 1}{\bold h_i(\theta)}\nabla_{\theta}\bold h_i(\theta) - \frac{1}{2}\sum_{i=1}^V (\bold U^T \bold s)_i^2 \nabla \bold h_i(\theta)
\end{flalign*}

Parameterised in this fashion, the distribution may be sampled from as:
\begin{flalign*}
  & \bold s =  \bold U \bold D(\theta)^{-1/2} \bold z 
\end{flalign*}
When $\bold z \sim \mathcal N(\bold 0, \bold I_V)$. Hence the time complexity of both density evaluation and sampling is determined by that of our linear operators.

\subsection{Gaussian Markov Random Fields}

Gaussian Markov random fields (GMRFs) provide a helpful framework for this enterprise (\citet{rue2005gmrfs}). They have been used, for example, to model spatial distributions in epidemiology (\citet{papageorgiou2014profile}) and in climate science, (\citet{zammit2015multivariate}). 

A GMRF is a multivariate Gaussian some of whose entries are independent of each other, conditional on the rest of the vector. This implies that if $\bold r \sim \mathcal N(\bold 0, \bold Q^{-1})$ is a GMRF, then $\bold Q$ has some level of sparsity where the precision matrix encodes conditional dependence properties.

Consider $\bold Q(\theta) = g(\theta) \bold I_V + h(\theta) \bold L$. Suppose the distribution models a graph in which the $l^{th}$ node has $\nu_l$ neighbours. Let $i\sim j$ if and only if $i$ and $j$ are neighbours, this information can be encoded in the Laplacian matrix $\bold L$ as:
\begin{flalign*}
  & \bold L_{i,j} = \begin{cases} 
    \nu_i & i=j \\ 
    -1 & i \sim j \\
    0 & \text{otherwise}
  \end{cases}
\end{flalign*}
This format can be used with respect to an arbitrary structure of neighbourhoods.

A result stated in \citet{merris1994laplacian} is that the Laplacian matrix of the Cartesian product of two graphs is the Kronecker sum of those graphs' Laplacian matrices. A useful implication for our present purposes is that if $\bold L_1, \bold L_2$ are two 1D Laplacian matrices of rank $N_1, N_2$ respectively with eigendecompositions $\bold U_i \bold D_i \bold U_{i}^T $ then we can find the eigendecomposition of the Cartesian product graph Laplacian as:
\begin{flalign}\label{eqn:DCT2D}
  & \bold L_{(2D)} = \bold L_1 \oplus \bold L_2 \\ \nonumber
  & = (\bold U_1 \otimes \bold U_2)(\bold D_1 \otimes \bold I_{N_2} + \bold I_{N_1} \otimes \bold D_2 )(\bold U_1 \otimes \bold U_2)^T \\ \nonumber
  & = \bold U_{(2D)} \bold D_{(2D)} \bold U_{(2D)T}
\end{flalign}
The 3D case is then the Kronecker sum of a 1D case and a 2D case, hence if $\bold L_3$ plays a natural role we find that the eigendecomposition of the 3D case is:
\begin{flalign*}
  & \bold L_{(3D)} = \bold L_1 \oplus \bold L_2 \oplus \bold L_3 = \bold U_{(3D)} \bold D_{(3D)} \bold U_{(3D)}^T\\ 
\end{flalign*}
Where 
\begin{flalign}\label{eqn:DCT3D}
  & \bold U_{(3D)} = \bold U_1 \otimes \bold U_2 \otimes \bold U_3 \\ \nonumber
  & \bold D_{(3D)} = \bold D_1 \otimes \bold I_{N_2} \otimes \bold I_{N_3}  + \bold I_{N_1} \otimes \bold D_2 \otimes \bold I_{N_3} \\ \nonumber
  & + \bold I_{N_1} \otimes \bold I_{N_2} \otimes \bold D_{3}
\end{flalign}

This approach allows us to break a complex modelling problem into elements for which closed form solutions exist, or for which numerical methods are feasible. We next illustrate this in the case of an adjacency model on a regular $n-$dimensional grid.

\subsection{Example: a graph for smooth features}
\label{sec:dctmodel}

Consider a graph on a line such that each node $\bold s_i$ is a neighbour of $\bold s_{i+1}$ and $\bold s_{i-1}$ wherever those indices are valid. That is $\bold s$ is a form of first order autoregressive process. Since $i \sim j$ if and only if $|i-j|=1$ we have that $\bold L$ is tridiagonal and thus sparse.  

\begin{figure}[htp]
  \centering
  \includegraphics[width=.5\textwidth]{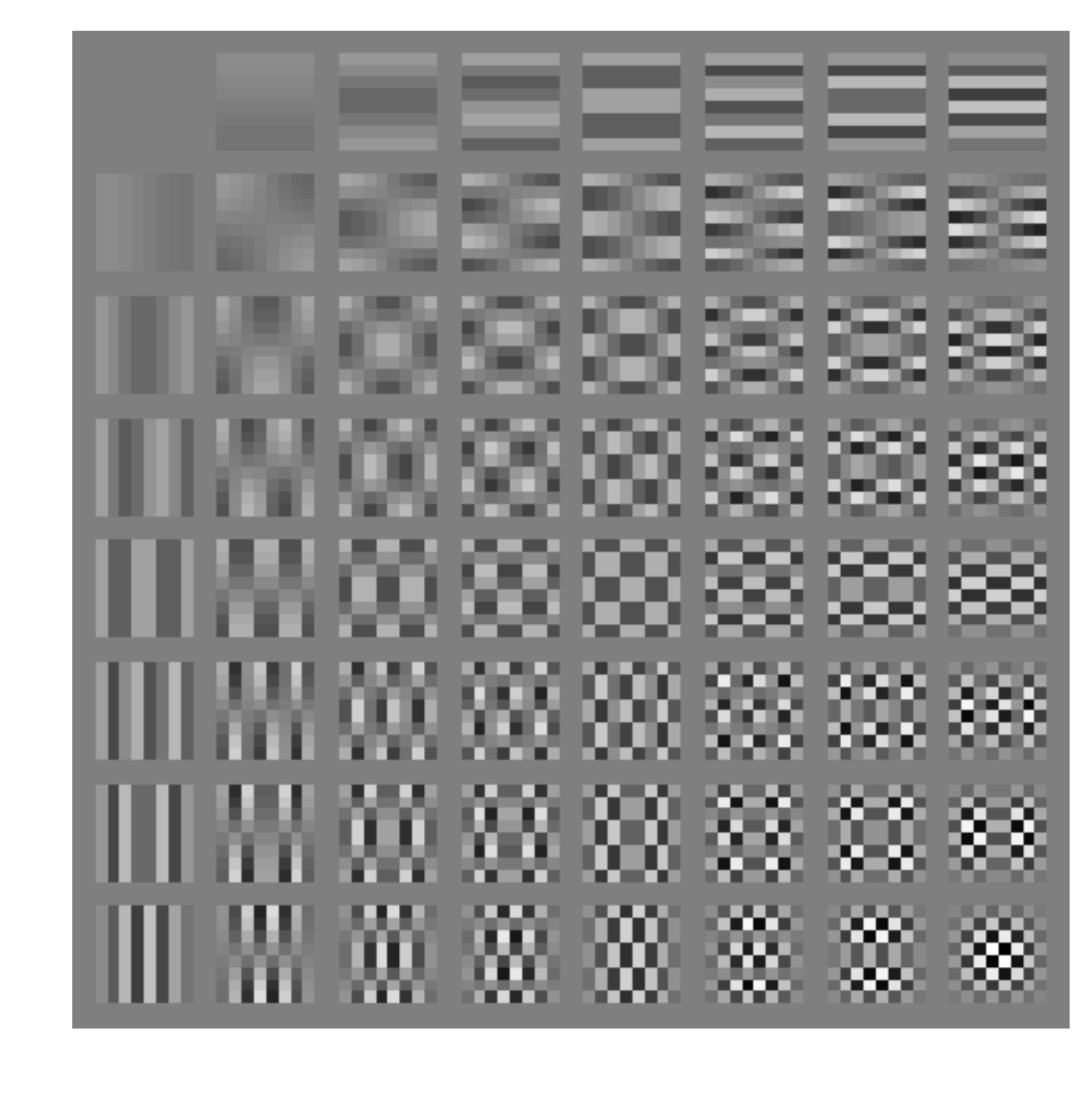}
  \caption{Scaled eigenvectors ($\bold U \bold D^{1/2}$) of $\bold L_{(2D)}$ on an $8 \times 8$ grid. Random vectors $\bold s$ which place large weights on `high-frequency' patterns will be penalised.}
  \label{fig:Qeigenvecs}
\end{figure}

We can obtain an analytical expression for the eigendomposition ($\bold L = \bold U \bold D \bold U^T$) of this Laplacian as the matrix encodes the recurrence requirements of the Discrete Cosine Transform (DCT). The analytical form of DCT-II is known (\citet{strang1999discrete}, \citet{ahmed1974dct}) and therefore (with $j,k$ ranging over $0,...,N-1$):
\begin{flalign}\label{eqn:DCTentries} 
  & \bold U_{jk} = \cos \left( (j + 1/2) k \pi / N \right) \\ \nonumber
  & \bold \gamma_k = 2 - 2\cos(k \pi / N)
\end{flalign}
This can be integrated into our framework with $\bold h_i(\theta) = \theta_1 + \theta_2 \gamma_i$ for fixed $\gamma_i, i=1,\ldots,V$, giving:
\begin{flalign*}
  & \bold Q(\theta) = \theta_1 \bold I_V + \theta_2\bold U \text{diag}(\{ \gamma_i \}) \bold U^T 
\end{flalign*}
We can extend this decomposition to 2- and 3- dimensional grids using either (\ref{eqn:DCT2D}) or (\ref{eqn:DCT3D}) respectively. These observations afford further insight into the random vector $\bold s \sim \mathcal N(\bold 0, \bold Q(\theta)^-1)$ as
\begin{flalign*}
  & \ln f(\bold s | \theta) \propto c + \theta_1 \bold s^T \bold s +  \theta_2 \sum_{i=1}^V \gamma_i (\bold U^T \bold s)_i^2
\end{flalign*}
from which we see that the distribution penalises by a flexible amount, $\theta_2$, those DCT-II coefficients, $(\bold U^T \bold s)_i^2$, corresponding to large eigenvalues, $\gamma_i$. Those indices $i$ for which the eigenvalues $\gamma_i$ is large are exactly those for which the corresponding eigenvectors represent high frequency activity, as illustrated in Fig \ref{fig:Qeigenvecs}.

\subsection{Other models}

\citet{rue2005gmrfs} discuss the link between another graphical model and discrete linear transform. They showed that for circulant block precision matrices the relevant transform is the Discrete Fourier Transform (DFT). The difference between DFT and DCT-II being the relevant boundary conditions where DFT essentially considers graphs on a loop or torus, while DCT-II has free edges. While that work considered graphs on regular grids it did not make fully clear the transform interpretation or the implication in terms of penalisation.

Another popular linear transform that yields an interpretable graphical model is the Discrete Wavelet Transform (DWT). Suppose the IDWT provides the eigenvectors $\bold U$ of a precision matrix for a particular orthogonal wavelet basis. A simple example would be a model for a fully connected graph. Here the discrete Haar or D4 wavelet bases, combined with the eigenvalue $0$ for the eigenvector $\bold 1$, as usual, and the eigenvalue $n$ for all other eigenvectors produce the required Laplacian matrix, $n\bold I - \bold 1 \bold 1^T$.

\begin{tabular}{ |p{1.3cm}|p{2.2cm}|p{3.3cm}|  }
 \hline
 \multicolumn{3}{|c|}{ } \\
 \multicolumn{3}{|c|}{ Table 2.1: collection of latent covariance structures} \\
 \hline
 \hline
 $\bold U$ & complexity & graphical model  \\ 
 \hline
 DCT & $\mathcal O(n \ln n)$ & autocorrelation \\  
 \hline
 DCT $\otimes$ DCT & $\mathcal O(n' \ln n')$, $n'=\max (n_1,n_2)$& spatial autocorrelation \\  
 \hline
 DFT & $\mathcal O(n \ln n)$  & circulant autocorrelation \\   
 \hline
 DWT-Haar & $\mathcal O(n)$ & fully connected groups \\
 \hline
 DWT $\otimes$ DCT & $\mathcal O(\max \{ n_1,$ $ n_2 \log n_2 \} )$ & fully connected groups autocorrelated over time \\
 \hline
 Any & $\mathcal O( n^3 )$ & aribitrary precision matrix \\
 \hline
 Any $\otimes$ DCT & $\mathcal O( \max \{n_1^3, $ $n_2 \log n_2 \} )$ & aribitrary precision matrix   autocorrelated over time\\
 \hline
\end{tabular}

\

The decompositions in (\ref{eqn:DCT2D}) and (\ref{eqn:DCT3D}) also help with Laplacians which are the product of characteristically different graphs. For instance suppose we were interested in modelling daily fluctuations of temperature at different spatial locations. Midnight being both the end and start of a day, $\bold L_1$ could represent a model where the daily timeseries was associated at its edges with a DFT based decomposition. While spatial dependency between nearby locations could be expressed with a 2D DCT based graph $\bold L_2 \oplus \bold L_3$. These models are summarised in Table 2.1.

A further set of rules for generating new covariance structures involves powers and sums of known structures.
\begin{flalign*}
  & \bold Q'(\theta) = \bold Q(\theta)^n + \bold Q(\theta)^m = \bold U(\bold D(\theta)^n + \bold D(\theta)^m)\bold U^T
\end{flalign*}
So that $\bold h'_i(\theta) = \bold h_i(\theta)^n + \bold h_i(\theta)^m$. For example, with the DCT based neighbourhood model and $n=2, m=0$, we find $\bold Q^2$ expresses second order relationships among the nodes. Similarly we can mix parameters:
\begin{flalign*}
  & \bold Q'(\theta') = \bold Q(\theta_{(1)}) + \bold Q(\theta_{(2)}) = \bold U(\bold D(\theta_{(1)}) + \bold D(\theta_{(2)}))\bold U^T
\end{flalign*}
So that $\bold h'_i(\theta) = \bold h'_i(\theta_{(1)}) + \bold h'_i(\theta_{(2)}) $.
These combinatorial rules for kronecker structure, powers, sums, and combination of spectral curves provide a powerful framework for working tractably with a range of nondiagonal latent covariance structures.

\section{Model} 
\label{sec:model}
 
In this section we propose an extension, infinite Sparse Structured Factor Analysis (iSSFA), to the work of \citet{knowles2007infinite}. Our model provides nondiagonal covariance structure for the latent features. The full specification of our model is:
  
\begin{center}
\begin{tabular}{ l l }
 $\bold Z \sim IBP(\alpha, \beta)$ & feature activations \\ 
 $\bold A_{t,j} \sim \mathcal N(\tau_j, \nu_j^{-1}) \ \ j=1,...$  & feature weights \\ 
 $\bold S_{j,:}^T \sim \mathcal N(\bold 0, \bold Q(\theta)^{-1}) \ \  \ j=1,... $ & features \\
 $\bold E_{t,:} \sim \mathcal N(\bold 0, \sigma^2 \bold I) \ \ \ \ t=1,...,T$ & measurement error \\
 $\sigma^2 \sim \mathcal{IG}(a,b)$ & noise level\\
 $\alpha \sim \mathcal{G}(e_\alpha,f_{\alpha})$ & IBP strength \\
 $\beta \sim \mathcal{G}(e_\beta, f_\beta)$ & IBP repulsion \\
 $\nu_j \sim \mathcal{G}(e_\nu, f_\nu)$ & weights' precisions \\
 $\tau_j \sim \mathcal N( m_\tau, r_\tau^{-1})$ & weights' means \\
 $\ln \theta_p \sim \mathcal{N}(m_p, r_p^{-1}), \ \ p=1,\ldots,P $ & feature parameters \\
\end{tabular}
\end{center}

Where we model $\bold Q(\theta) = g(\theta) \bold I_V + h(\theta) \bold L = \theta_1 \bold I_V + \theta_2 \bold L$. We will take $\bold L$ to be a multi-dimensional neighbourhood graph as discussed in Section \ref{sec:dctmodel} and hence made efficient through the DCT. In our simulated example of Section \ref{sec:simulation} this will be a 2D graph, while in neuroimaging example of Section \ref{sec:neuroexample} it will be 3D. However, the scheme provided below would work for other suitable $\bold Q$ as described in Section \ref{sec:textures} with little modification. Observe that the distribution of features requires that they be smooth to a level determined by the spatial hyperparameter $\theta$, which we then learn from the data. 

Compared to \citet{knowles2007infinite} the enriched prior on $\bold A$ allows different means and variances for each column. This adds expressivity to the model when sources are not identically distributed. Also, we observe that zero mean weights in $\bold A$ can lead more easily to degenerate situations with respect to sparsity in which $\bold Z = \bold 1_{T,K}$. We will see in Section \ref{sec:gibbs_Z_old} that non-zero means enable the relevant spatial vector $\bold S_{k,:}^T$ to explain residuals when deciding whether to activate a feature for an observation $\bold Z_{t,k}$.

\section{Gibbs sampler}
\label{sec:sampler}

Building on the methods of \citet{knowles2007infinite} we can construct a Gibbs sampler for the model proposed in Section \ref{sec:model}. The sampler is approximate in that we use a Laplace approximation to the posterior distribution of the latent feature parameters. 

For clarity, in our notation $\bold M_{i,:}$ means the $i^{th}$ row of a matrix, or, if $i$ is a set, all rows indexed by elements of the set. A minus sign in the index $\bold M_{-i,:}$ means all rows except the $i^{th}$, with a similar convention for sets. Similar conventions apply to indexing of columns.

\subsection{Conditional distributions of latent features, $\bold S_{k,:}^T$}

Consider resampling the feature $\bold S_{k,:}$. Define a residual:
\begin{flalign*}
  \boldsymbol \epsilon_t^k = (\bold Y_{t,:} - (\bold A_{t,-k} \circ \bold Z_{t,-k}) \bold S_{-k,:})^T
\end{flalign*}

The Gibbs conditional distribution for $\bold S_{k,:}^T$ is, then, multivariate normal

\begin{flalign*}
  & \bold S_{k,:}^T | \bold A, \bold Z, \bold S_{-k,:}, \sigma^2, \bold Y \sim \mathcal N(\boldsymbol \mu_{\bold S}, \boldsymbol \Sigma_{\bold S}) \\
  & \text{with} \\
  & \boldsymbol \mu_{\bold S} = \left(\frac{\sum_{t=1}^{T} \bold A_{t,k}^2\bold Z_{t,k} }{\sigma^2}\bold I + \bold Q\right)^{-1}
  \left(\frac{1}{\sigma^2}\sum_{t=1}^{T} \bold A_{t,k}\bold Z_{t,k} \boldsymbol \epsilon_t^k \right) \\
  & \boldsymbol \Sigma_{\bold S} = 
  \left(\frac{\sum_{t=1}^{T} \bold A_{t,k}^2\bold Z_{t,k} }{\sigma^2}\bold I + \bold Q\right)^{-1}
\end{flalign*}

The feature vectors function as a basis for the latent data space. We can see from these conditional distributions that the Gibbs sampler bears a resemblance to the Gram-Schmidt orthogonalisation procedure. Each feature is treated as though it were the final vector in the procedure, the effects of the other features having being subtracted from the data to leave a residual vector which is then normalised by application of a covariance matrix.

\subsection{Activation of shared features within $\bold Z_{t,k}$}
\label{sec:gibbs_Z_old}

Sampling of $\bold Z$ occurs in two phases. We describe features as shared when $m_k = \sum_{t=1}^T \bold Z_{t,k}>1$ and as unique when $m_k = 1$. A draw from the two parameter IBP is obtained by, for the $t^{th}$ observation, activating each shared feature with probability $\sum_{r=1}^{t-1} \bold Z_{r,k} / (\beta + t -1)$ and then drawing a set of $\text{Poisson}(\alpha \beta / (\beta + t - 1))$ unique features (\citet{knowles2007infinite}). The rows of the matrix are exchangeable so that conditional on the other observations, the $t^{th}$ observation may be treated as the $T^{th}$. When doing posterior inference, this structure leads naturally to two stages, in which we sample shared features as Bernoulli, and a number of unique features as Poisson. The following step deals with resampling of shared features.

For $\bold Z$ we will approach sampling entry-wise, using the technique of \citet{knowles2007infinite} in defining a ratio of conditionals $r = r_l r_p$. Where $r_p$ is available from section 2.1 of that work, $r_p = \frac{m_{k,-t}}{\beta + T -1 - m_{k,-t}}$, with $m_{k,-t}$ being the number of times the $k^{th}$ factor has been activated, excluding observation $t$. In our case $r_l$ is different with:

\begin{flalign*}
  & r_l = \frac{ p(\bold Y_{t,:} | \bold A_{:,-k}, \bold A_{-t,k}, \bold S, \bold Z_{:,-k}, \bold Z_{-t,k}, \bold Z_{t,k}=1,\sigma^2) }
  {p(\bold Y_{t,:} | \bold A_{:,-k}, \bold A_{-t,k}, \bold S, \bold Z_{:,-k}, \bold Z_{-t,k}, \bold Z_{t,k}=0,\sigma^2 )}
\end{flalign*}

The required marginal distribution of the numerator is again Gaussian:
\begin{flalign*}
  & \bold Y_{t,:}^T | \bold A_{:,-k}, \bold A_{-t,k}, \bold S, \bold Z_{:,-k}, \bold Z_{-t,k}, \bold Z_{t,k}=1,\sigma^2, \cdot \sim \\
  & \mathcal N \left( \left[ (\bold A_{t,-k} \circ \bold Z_{t,-k})\bold S_{-k,:} \right]^T + \bold S_{k,:}^T \tau_k, \ \
  \sigma^2 \bold I_V + \frac{1}{\nu_k} \bold S_{k,:}^T \bold S_{k,:} \right)
\end{flalign*}
This presents an important modelling difference from the work in \citet{knowles2007infinite} where, due to the assumption of zero-mean feature weights ($\tau_k = 0$), the mean of the numerator distribution would not feature $\bold S_{k,:}^T$. That fact would force the algorithm to explain the residual $(\bold Y_{t,:} - (\bold A_{t,-k} \circ \bold Z_{t,-k})\bold S_{-k,:} )^T$ using only the covariance structure. Non-zero means $\tau_k$ allows a feature vector to help explain the residual when sampling a feature activation $\bold Z_{t,k}$. Now set:

\begin{flalign*}
  & \bold r_0 = \bold Y_{t,:}^T - \left[ (\bold A_{t,-k} \circ \bold Z_{t,-k})\bold S_{-k,:} \right]^T\\
  & \bold r_1 = \bold Y_{t,:}^T - \left[ (\bold A_{t,-k} \circ \bold Z_{t,-k})\bold S_{-k,:} \right]^T - \bold S_{k,:}^T \tau_k
  \end{flalign*}
  
We can now evaluate $\ln r_l$ as:
\begin{flalign*}
  & \ln r_l = \ln p(\bold Y_{t,:}| \bold Z_{t,k} = 1, \cdot) - \ln p(\bold Y_{t,:}| \bold Z_{t,k} = 0, \cdot) = \\
  & \ \ \ -\frac{1}{2} \ln \left(1 + \frac{1}{\sigma^2 \nu_k} \bold S_{k,:}\bold S_{k,:}^T \right) - \frac{1}{2\sigma^2} \bold r_1^T \bold r_1 \\
  & \ \ \ + \frac{1}{2}((\sigma^2 \nu_k)(\sigma^2 + \bold S_{k,:}\bold S_{k,:}^T/\nu_k ))^{-1} (\bold S_{k,:}\bold r_1)^2 \\
  & \ \ \ + \frac{1}{2\sigma^2} \bold r_0^T \bold r_0
\end{flalign*}
Hence, despite both the precision and covariance matrices for the numerator being dense, the density may be evaluated in $\mathcal O(V)$ time and memory as the main calculations are dot products of dense vectors in $\mathbb R^V$. For the shared features, $k$, we can now sample $\bold Z_{t,k} \sim \text{Bern}\left(\frac{r_lr_p}{r_lr_p + 1}\right)=\text{Bern}((1+\exp(-\ln(r_l r_p)))^{-1})$.

We use the techniques described in \citet{pearce2016scaling} to efficiently parallelise these calculations across a compute cluster.

\subsection{Feature weights for shared features, $\bold A_{t,k}$}

Setting $\mu_t = [(\bold A_{t,-k} \circ \bold Z_{t,-k})\bold S_{-k,:}]^T$, the Gibbs conditional for $\bold A_{t,k}$ is Gaussian with:

\begin{flalign*}
  & \bold  A_{t,k} | \cdot \sim \mathcal N(\mu_A, \zeta_A) \\
  & \text{with} \\
  & \mu_A = \left(\nu_k + \frac{\bold S_{k,:}\bold S_{k,:}^T}{\sigma^2} \right)^{-1}
  	\left( \frac{1}{\sigma^2} \bold S_{k,:}(\bold Y_{t,:}^T - \mu_t) + \nu_k \tau_k \right) \\
  & \zeta_A = \left(\nu_k + \frac{\bold S_{k,:}\bold S_{k,:}^T}{\sigma^2} \right)^{-1} 
\end{flalign*}

\subsection{Activating unique features}
\label{ssec:ZAunique}

In order to complete a draw from the Indian Buffet Process we must sample unique features for each observation, a task we will tackle by a Metropolis-Hastings step for each observation. The IBP is exchangeable (\citet{griffiths2011indian}, \citet{knowles2007infinite}) hence the $t^{th}$ observation may be treated as the $T^{th}$. Define $J_t$ to be the set of indices of $n$ new components for observation $t$. We set the number of such components and the parameters and weights associated with them as $\omega$ = $( n, \{\tau_j\}, \{\nu_j\}, \{a_j\} )$ where the $a_j$ are weights for features.

We make a proposal $\omega^*$ in three steps: sampling the number of new sources $n$, then their parameters $\{\tau_j\}, \{\nu_j\} | n $ and finally the weightings $ a_j | \{\tau_j\}, \{\nu_j\}$. Note that at this point the feature vectors will be marginalised out. We then choose to accept or reject the proposal via an MH step.  It will be convenient to write, 
\begin{flalign*}
  & \boldsymbol \mu = \bold S_{-J_t,:}^T(\bold A_{t,-J_t} \circ \bold Z_{t,-J_t})^T 
\end{flalign*}
 as the values of the columns of $\bold A, \bold Z$, and rows of $\bold S$ corresponding to shared features enter in to our calculations only through the mean of $\bold Y_t$. Afterwards, conditional on MH acceptance, we will sample the new sources $\{\bold s_j\}_{j \in J_t}$ (see Section \ref{ssec:Sunique}). That is to say we block sample $p(\omega, \{\bold s_j \} | \cdot)$ as $p(\{\bold s_j \} | \omega, \cdot )p(\omega | \cdot)$ for each $t$.

To begin, we use the respective priors to generate proposals:
\begin{flalign*}
  & n \sim Poi \left(\frac{\beta \alpha}{\beta + T - 1} \right) \\
  & \tau_j | n \sim \mathcal N(m_\nu, r_\nu^{-1}) \ , \ \ j \in J_t \\
  & \nu_j | n \sim \mathcal G(e_\nu, f_\nu) \ , \ \ j \in J_t \\
  & a_j | n, \tau_j, \nu_j \sim \mathcal N(\tau_j, \nu_j^{-1}) \ , \ \ j \in J_t
\end{flalign*}
Then form an acceptance probability ratio where we make use of the proposal $q(\omega^*| \omega)$ being equal to the prior $g(\omega^*)$ so that:
\begin{flalign*}
 & \psi = \frac{p( \omega^* | \bold Y_{t,:}^T, \cdot)q( \omega | \omega^*)}
 {p(\omega | \bold Y_{t,:}^T, \cdot)q(\omega^* | \omega )} = 
 \frac{p( \bold Y_{t,:}^T |\omega^*,\cdot)}{ p( \bold Y_{t,:}^T |  \omega, \cdot )}
\end{flalign*}
Which forms the basis of the acceptance probability. 


We now focus on the density for $f( \bold Y_{t,:}^T | \omega )$, for which we need to marginalise out the features corresponding to the weights we sampled. In their prior, the features $\{\bold s_j\}$ are, conditional on the hyperparameters, IID multivariate normal. Let us stack them into a single vector, and define a block matrix which we will use to manipulate them.

\begin{flalign*}
 & \bold r = \left[ \begin{matrix}
 \bold s_1 \\
 \bold s_2 \\
 \vdots \\
 \bold s_{n}
 \end{matrix} \right] \ \ , \
 \bold C = \left [ a_1 \bold I_V, a_2 \bold I_V, \cdots, a_n \bold I_V\right]
\end{flalign*}

So that $\bold r$ is $(nV \times 1)$ and $\bold C$ is $(V \times nV)$. The conditional distribution of the observations is, then:

\begin{flalign}\label{eqn:kronunique}
  & \bold Y_{t,:}^T | \{ a_j \}, \bold r, \cdot 
  	\sim \mathcal N_V\left( \boldsymbol \mu + \bold C \bold r, \sigma^2 \bold I_V \right) \nonumber \\
 & \bold r \sim \mathcal N_{nV}(\bold 0, \bold I_n \otimes \bold Q^{-1})
\end{flalign}

Hence, using results on the multivariate normal distribution:

\begin{flalign*}
 & \bold Y_{t,:}^T | \{a_j\}, \cdot \sim \mathcal N_V\left( \boldsymbol \mu , \ \ \sigma^2 \bold I + \left(\sum_{j \in J_t} a_j^2 \right) \bold Q^{-1} \right)
\end{flalign*}

This density function provides the numerator and denominator of $\psi$, enabling us to calculate the acceptance probability for the proposal.

\subsection{ Drawing spatial feature vectors for unique features }
\label{ssec:Sunique}

This section continues with the notation used in Section \ref{ssec:ZAunique}, as we are completing a block draw. For each observation $t$ we need to draw the spatial features from the distribution $\{\bold s_j \} | \omega, \cdot$. Hence, reversing the conditioning in (\ref{eqn:kronunique}) we see that the posterior for $\bold r$ is again Gaussian:

\begin{flalign*}
 & \bold r | \bold Y_{t,:}^T, \cdot \sim \mathcal N(\boldsymbol \mu_\bold r, \boldsymbol \Sigma_\bold r) \\
 & \boldsymbol \mu_\bold r = \left(\frac{1}{\sigma^2} \bold C^T \bold C + \bold I_n \otimes \bold Q \right)^{-1} \left( \frac{1}{\sigma^2} \bold C^T(\bold Y_{t,:}^T - \boldsymbol \mu) \right) \\
 & \boldsymbol \Sigma_\bold r =  \left(\frac{1}{\sigma^2} \bold C^T \bold C + \bold I_n \otimes \bold Q \right)^{-1} 
\end{flalign*}

\subsection{Sampling the noise level $\sigma^2$}

If we place an inverse gamma prior on $\sigma^2 \sim \mathcal{IG}(u,v)$ then conditional on the other variables, the result is conjugate. Write $\boldsymbol \epsilon_t = \bold Y_{t,:} - (\bold A_{t,:} \circ \bold Z_{t,:}) \bold S$ and note that there are $T$ such rows. The conditional follows an inverse gamma distribution:
\begin{flalign*}
 & \sigma^2 | \bold Y, \cdot \sim \mathcal IG\left(\frac{TV}{2}+u, v + \frac{1}{2} \sum_{t=1}^T \sum_{v=1}^V \boldsymbol \epsilon_{t,v}^{2}\right)
\end{flalign*}

\subsection{Sampling spatial parameters $\theta$}

As the paramaters of the latent feature must be non-negative we place our prior on their logarithms $\xi_p = \ln \theta_p$ for $p=1,\ldots,P$. We sample from the conditional posterior for $\xi$ using a Laplace approximation (\citet{rue2009approximate}) to the posterior distribution, that is to:
\begin{flalign*}
  & p(\xi | \bold S, \cdot) \propto p( \bold S| \xi, \cdot) f(\xi)
\end{flalign*}
We find that this approach works well across scales of $\theta$ without the need for additional MCMC tuning parameters required by asymptotically exact methods. 
For the prior we will assume independent normal distributions for the $\xi_p$ so that $f(\xi)=\prod_{p=1}^P \mathcal N(\xi_2; m_{\xi_2}, r_{\xi_2})$. 


To perform the Laplace approximation we must find the MAP estimate of the parameters. In our example model $\bold h_i(\theta) = \theta_1 + \theta_2 \gamma_i$ so that $\bold Q = \theta_1 \bold I_V + \theta_2 \bold L$. The log posterior is:
\begin{flalign*}
 & \ln f(\xi | \bold S, \cdot) = \ln f(\xi | \bold S) = c + \frac{K}{2} \sum_{i=1}^V \ln(\theta_1 + \theta_2 \gamma_i) \\
 & \ \ \  - \frac{1}{2} \sum_{i=1}^{K} \bold S_{i,:}(\theta_1 \bold I + \theta_2 \bold L)\bold S_{i,:}^T \\
 & \ \ \ - \frac{r_{\xi_1}}{2} \xi_1^2 + m_{\xi_1} r_{\xi_1} \xi_1 
 - \frac{r_{\xi_2}}{2} \xi_2^2 + m_{\xi_2} r_{\xi_2} \xi_2 
 - \xi_1 - \xi_2 
\end{flalign*}

Which, after applying the chain rule, provides the following first derivatives:

\begin{flalign*}
 & \frac{\partial \ln p(\xi | \bold S)}{\partial \xi_1} = 
 \theta_1 \left(\frac{K}{2} \sum_{i=1}^V \frac{1}{\theta_1 + \theta_2 \gamma_i} 
 - \frac{1}{2} \sum_{i=1}^{K} \bold S_{i,:} \bold S_{i,:}^T \right) \\
 & \ \ \ - r_{\xi_1} \xi_1 + m_{\xi_1} r_{\xi_1} - 1  \\ 
 & \frac{\partial \ln f(\xi | \bold S)}{\partial \xi_2} = 
 \frac{K}{2} \sum_{i=1}^V \frac{\gamma_i}{\theta_1 + \theta_2 \gamma_i} 
 - \frac{1}{2} \sum_{i=1}^{K} \bold S_{i,:} \bold L \bold S_{i,:}^T \\
 & \ \ \ - r_{\xi_2} \xi_2 + m_{\xi_2} r_{\xi_2} - 1  \\  
\end{flalign*}

After which the second derivatives required to populate the negative of the Hessian matrix, $\bold H$, can also be found:

\begin{flalign*}
 & \frac{\partial^2 \ln p(\xi | \bold S)}{\partial \xi_1^2} = 
  \frac{\partial \ln p(\xi | \bold S)}{\partial \xi_1} 
  -\theta_1^2 \frac{K}{2} \sum_{i=1}^V (\theta_1 + \theta_2 \gamma_i)^{-2}
  -r_{\xi_1}  \\ 
 & \frac{\partial^2 \ln p(\xi | \bold S)}{\partial \xi_2^2} = 
 \frac{\partial \ln p(\xi | \bold S)}{\partial \xi_2}
 - \theta_2^2 \frac{K}{2} \sum_{i=1}^V \left( \frac{\gamma_i}{\theta_1 + \theta_2 \gamma_i} \right)^2
  - r_{\xi_2} \\  
 & \frac{\partial^2 \ln p(\xi | \bold S)}{\partial \xi_1 \xi_2} = -\theta_1 \theta_2 \frac{K}{2} \sum_{i=1}^V \frac{\gamma_i}{\left(\theta_1 + \theta_2 \gamma_i \right)^2}
\end{flalign*}

We then use numerical optimisation to find a MAP estimate $\hat \xi$, following which we sample from the approximate conditional posterior distribution:
\begin{flalign*}
 & \xi | \bold S, \cdot \sim \mathcal N(\hat \xi, \bold H^{-1})
\end{flalign*}
If required, this distribution can be used as a proposal for a corrected MH accept/reject step:
\begin{flalign*}
 & \frac{ f(\xi' | \bold S, \cdot) N(\xi; \hat \xi, \bold H^{-1})}{f(\xi | \bold S, \cdot) N(\xi' ; \hat \xi, \bold H^{-1})} 
\end{flalign*}
Which, if the approximation is good should have a high acceptance rate, and where all the quantities can be evaluated using previous computations. However, in the examples below we sample from the Laplace approximation directly.

\subsection{Sampling the IBP strength $\alpha$}

We use a rate parameterised alternative to that in \citet{knowles2007infinite}. Let $H_T (\beta) = \sum_{i=1}^T \frac{\beta}{i + \beta - 1}$, then:
\begin{flalign*}
 & P(\alpha | \bold Z, \beta) \propto P \left( \bold Z | \alpha, \beta \right)P(\alpha) \\  
 & \propto \mathcal G(\alpha ; K_+ + e_\alpha, f_\alpha +  H_T(\beta) )
\end{flalign*}

Where $K_+$ is the number of non-zero columns of $\bold Z$.

\subsection{Sampling the IBP repulsion $\beta$}

We will sample this via a metropolis step. We use the shape-rate parametrised Gamma prior $\beta \sim \mathcal G(e_\beta, f_\beta)$. In order to ensure non-negativity we will work with parameter logarithms. 
We take a Normal distribution for the proposal $q( \psi | \psi^*) = \mathcal N( \psi^*; \psi, s_\beta)$. 
Write $m_k = \sum_{t=1}^T \bold Z_{t,k}$. $B(.,.)$ is the beta function, $H_T(\beta)$ is as above, and $K_l$ is the number of columns whose entries correspond to the integer $l$ expressed in binary (reading a column of $\bold Z$ downwards yields a binary number $0\le l \le 2^{T}-1$). The likelihood ratio is:

\begin{flalign*}
 & L = \ln \frac{P(\bold Z | \alpha, \psi^*)}{P(\bold Z | \alpha, \psi)} = - \alpha H_T(\beta^*) + \alpha H_T(\beta) \\
 & = \sum_{k=1}^{K_+} \left[ \ln \Gamma( T-m_k + \beta^*) - \ln \Gamma( T-m_k + \beta)  \right] \\ 
 & \ \ \ + K_+ (\ln \Gamma( T + \beta) - \ln \Gamma( T + \beta^*)) \\
 & \ \ \ + \alpha (H_T(\beta) - H_T(\beta^*))
\end{flalign*}

So that the log Metropolis-Hastings ratio is found as:
\begin{flalign*}
 & \ln a_{\beta \rightarrow \beta^*} = L + e_\beta \ln \beta^* - f_\beta \beta^* - e_\beta \ln \beta + f_\beta \beta \\
 &  = L + e_\beta[\ln \beta^* - \ln \beta] +  f_\beta[\beta -  \beta^*]
\end{flalign*}

\subsection{Sampling weight precisions $\nu_k$}
The Gibbs distibution for $\nu_k$ is a Gamma:
\begin{flalign*}
 & \nu_k | \{ \bold A_{:,k} \}_{t:Z_{t,k}=1}, \cdot \sim \mathcal G (\tilde e_\nu, \tilde f_\nu) \\
 & \text{with} \\
 & \tilde e_\nu = e_\nu + 1/2 \sum_{i=1}^T \bold Z_{t,k} , \\ 
 & \tilde f_\nu = \ f_\nu + \frac{1}{2}\sum_{i=1}^T ((\bold A_{t,k}-\tau)\bold Z_{t,k})^2
\end{flalign*}

\subsection{Sampling the weight means $\tau_k$}

The Gibbs distribution 
$\tau_k | \bold A, \bold Z, \cdot$ is normal with:

\begin{flalign*}
 & \tau_k | \bold A, \bold Z, \cdot \sim \mathcal N(\mu_\tau, \zeta_\tau) \\
 & \mu_\tau = \left(\nu_k \sum_{t=1}^T\bold Z_{t,k} + r \right)^{-1} \left( \nu_k \sum_{t:Z_{t,k}=1} \bold A_{t,k} + r m \right) \\
 & \zeta_\tau = \left(\nu_k \sum_{t=1}^T\bold Z_{t,k} + r \right)^{-1} 
\end{flalign*}

\section{Simulated Example}
\label{sec:simulation}

\begin{figure*}[htp]
  \begin{subfigure}{1.0\textwidth}
   \includegraphics[width=1.00\textwidth]{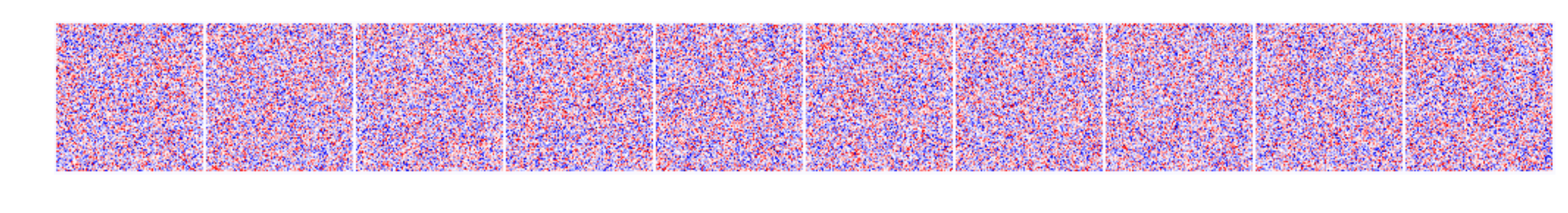}
   \caption{Observed images $\bold Y_{1:10,:}$ from holdout data}
   \label{fig:Y_obs}
  \end{subfigure}
  \begin{subfigure}{1.0\textwidth}
   \includegraphics[width=1.00\textwidth]{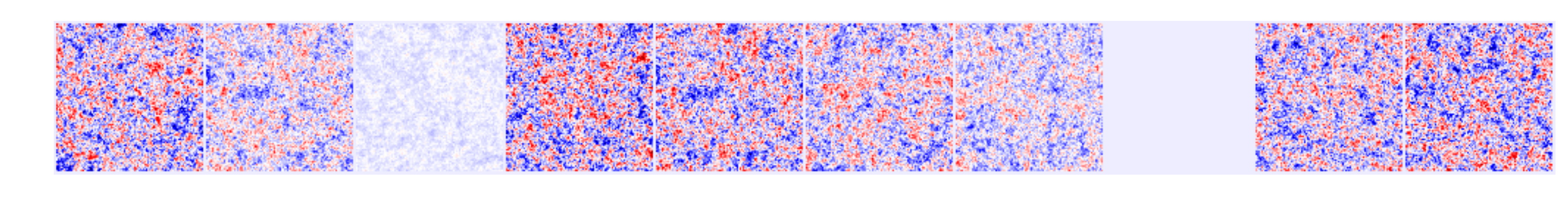}
   \caption{Corresponding latent vectors $(\bold A \circ \bold Z)_{1:10,:}\bold S$ }
   \label{fig:X_true}
  \end{subfigure}
  \begin{subfigure}{1.0\textwidth}
   \includegraphics[width=1.00\textwidth]{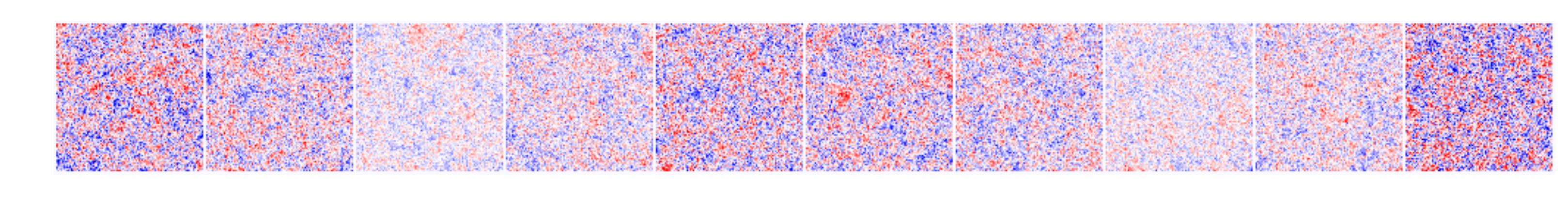}
   \caption{Corresponding PCA reconstruction with $K^*=33$}
   \label{fig:X_eig}
  \end{subfigure}
  \begin{subfigure}{1.0\textwidth}
   \includegraphics[width=1.0\textwidth]{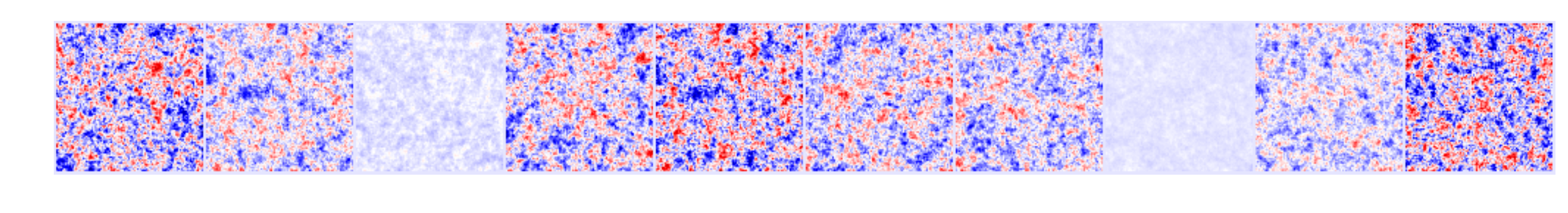}
   \caption{Corresponding Issfa reconstruction $\hat{\mathbb E}[(\bold A \circ \bold Z)_{1:10,:}\bold S]$}
   \label{fig:X_issfa}
  \end{subfigure}
  \begin{subfigure}{1.0\textwidth}
   \includegraphics[width=1.0\textwidth]{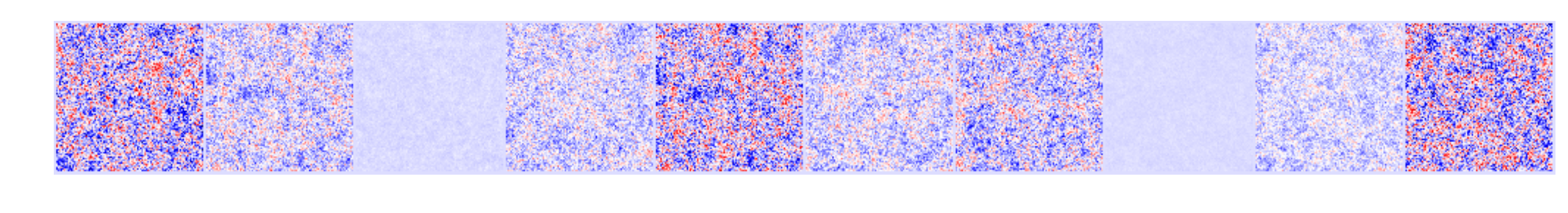}
   \caption{Similar to (d), but using spatially IID prior.}
   \label{fig:X_isfa}
  \end{subfigure}
  \caption{Different views of examples from the holdout dataset.}
  \label{fig:denoising}
\end{figure*}

We generated $T=3000$ observations on a $100\times 100$ grid ($V=10000$). Each observation was a linear combination of 50 features drawn as $\bold S_{k,:}^T \sim \mathcal N(\bold 0, \bold Q(1,100)^{-1})$ and then unit normalised.  Activation weights were drawn as $\bold Z_{t,k} \sim Bern(1/20)$ with $\bold A_{t,k}\sim \mathcal N(1,\gamma_k)$, $50<\gamma_k<100$, $\sigma^2 \sim \mathcal N(0, 1)$. Additive noise was drawn IID $\bold E_{t,k}\sim \mathcal N(0,1)$. A further holdout dataset of 500 images was drawn from the same distribution for performance testing. The first 10 observed images from the hold out data set are shown in Figure \ref{fig:Y_obs} and the corresponding latent images in Figure \ref{fig:X_true}. 

\begin{figure}[htp]
    \includegraphics[width=.45\textwidth]{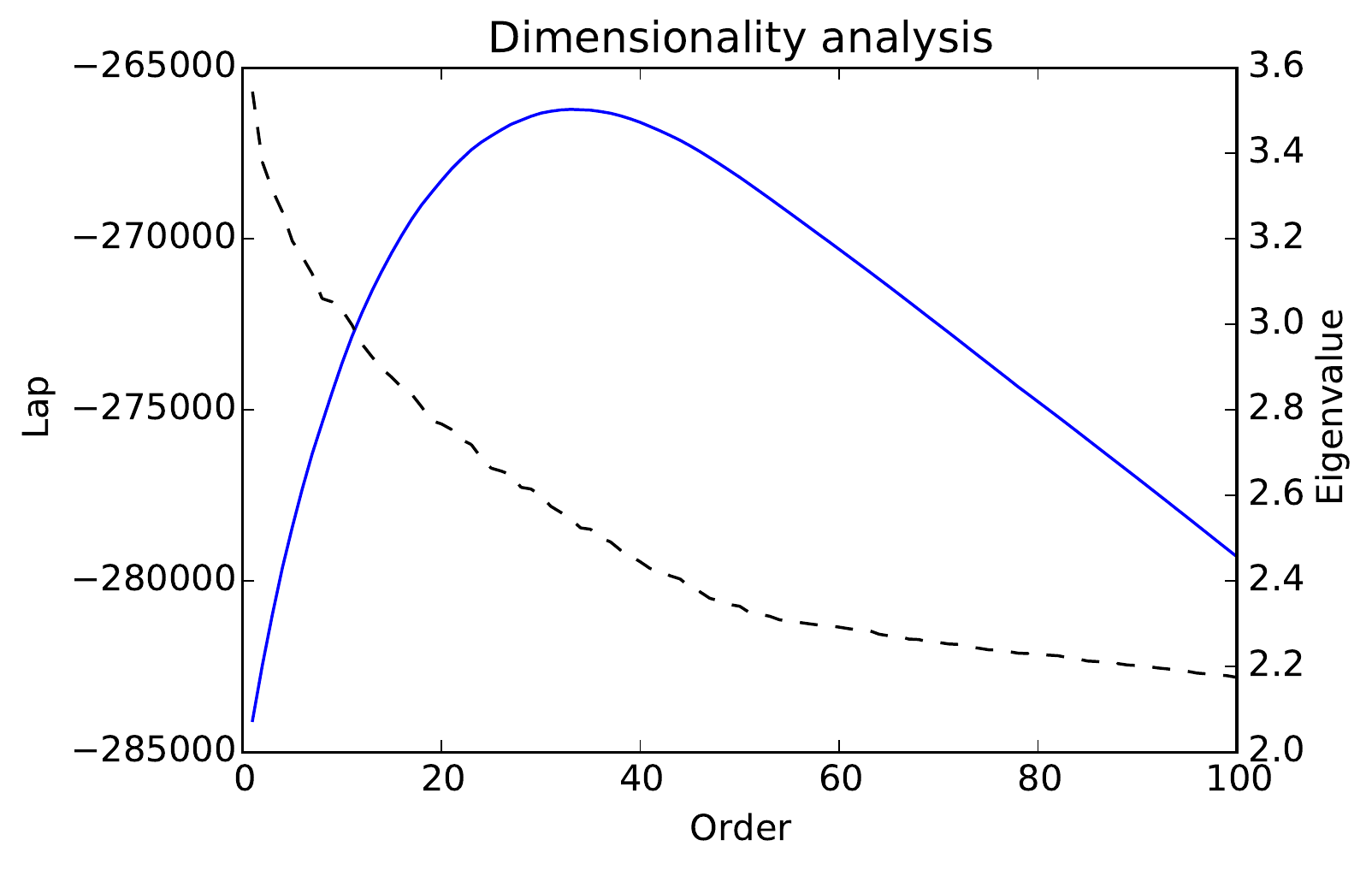}
    \caption{Analysis of the dimensionality of the dataset. Blue line shows the Laplace evidence approximation of the dimensionality obtained by Minka's method (optimising $K^*=33$). Black dashed line shows the first 100 eigenvalues of the data covariance matrix.}
    \label{fig:dimensionality}
\end{figure}

Six nodes of a high performance compute cluster were used to execute the analysis. Each node had twelve Intel Xeon E5-2620 2.00GHz CPUs and an Nvidia Tesla K20m GPU. The code was written in Julia (\citet{bezanson2014julia}) in order to utilise the multiprocessing capabilities of the cluster. Features for the sampler were initialised using a K-means algorithm fit for 15 clusters.

\begin{figure}[htp]
  \begin{subfigure}{0.5\textwidth}
    \includegraphics[width=1.0\textwidth]{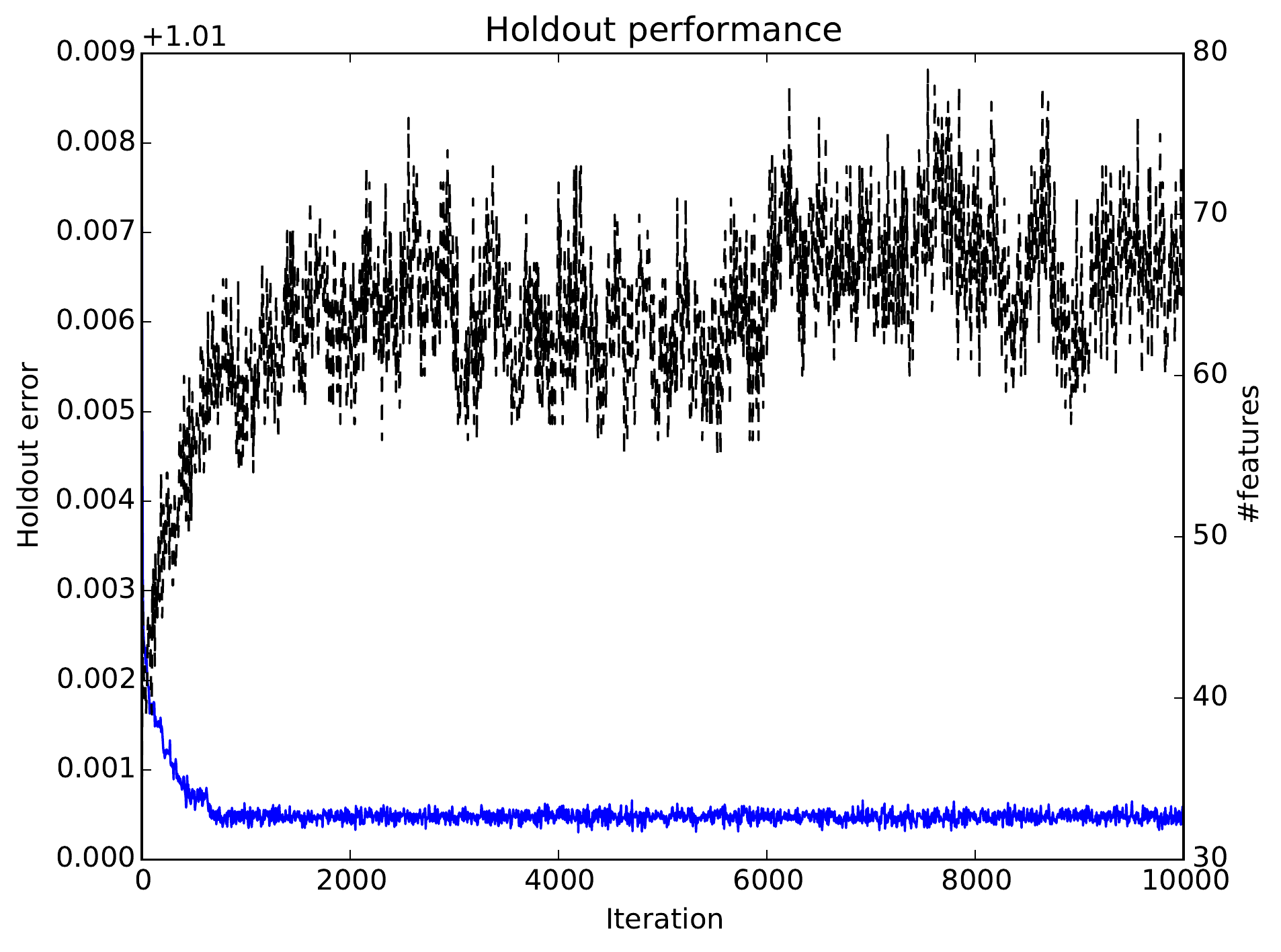}
    \caption{Blue line shows the reconstruction error of the $n^{th}$ MCMC sample on held out data. Black line shows the number of features available.}
    \label{fig:traceissfa}
  \end{subfigure}
  \begin{subfigure}{0.5\textwidth}
    \includegraphics[width=1.0\textwidth]{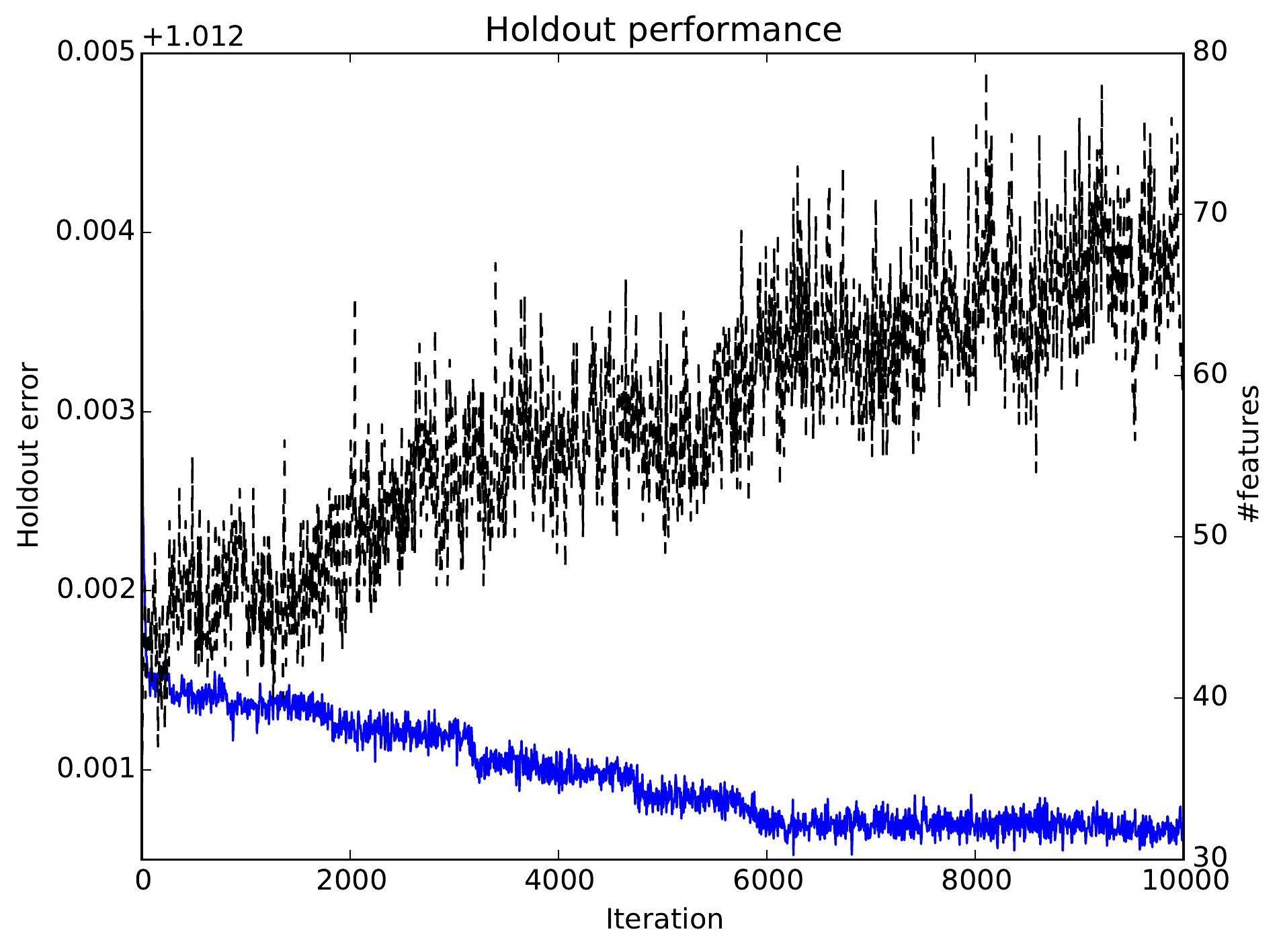}
    \caption{Same as (a), but using spatially IID prior of \citet{knowles2007infinite}.}
    \label{fig:traceissfa}
  \end{subfigure}
  \caption{MCMC sampling traces for analysis of the same dataset under different priors.}
  \label{fig:traces}
\end{figure}

Call the true latent data $\bold X = (\bold A \circ \bold Z)\bold S$. Figure \ref{fig:X_issfa} shows the mean reconstruction of the latent data. That is the Monte Carlo approximation: 
\begin{flalign*}
  & \mathbb E[(\bold A \circ \bold Z)\bold S] \approx \hat{\bold X}^{iSSFA} = \frac{1}{N} \sum_{n=1}^N (\bold A^{(n)} \circ \bold Z^{(n)})\bold S^{(n)}
\end{flalign*}
where superscripts $(n)$ indicate samples from particular MCMC iterations. Figure \ref{fig:X_isfa} shows the same estimate under the spatially IID prior of \cite{knowles2007infinite}. In both cases we used a thinned subsequence of the MCMC samples to obtain the estimate.

\begin{figure}[htp]
  \begin{subfigure}{0.5\textwidth}
   \includegraphics[width=1.0\textwidth]{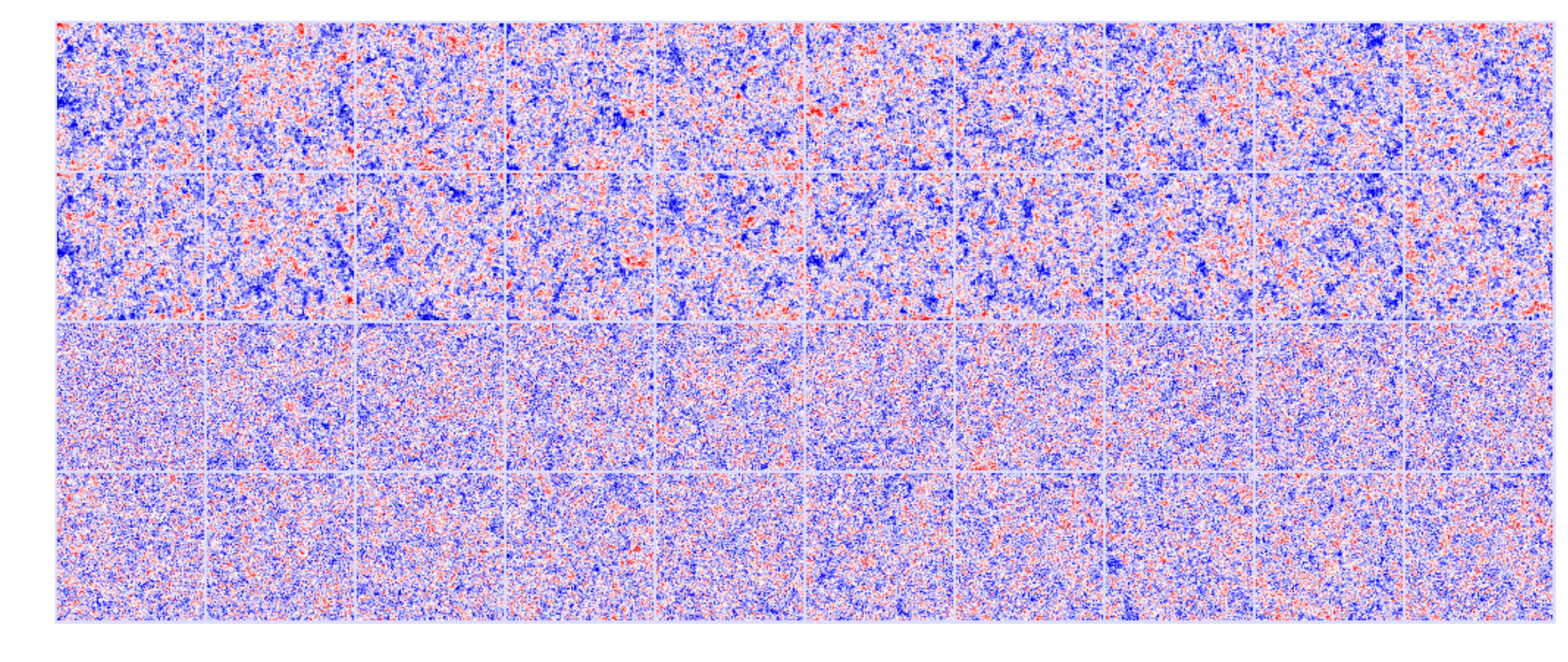}
  \caption{Qualitative comparison of feature recovery. Top row: first ten true features. Second row: matched iSSFA features (final MCMC sample); third row matched FastICA features; bottom row: matched eigenvectors. The eigenvectors and the related ICA features appear to vary on a dissimilar scale to the true features. The iSSFA features are all of a similar scale to the true features.}
  \label{fig:features}
  \end{subfigure}
  \begin{subfigure}{0.5\textwidth}
    \includegraphics[width=1.0\textwidth]{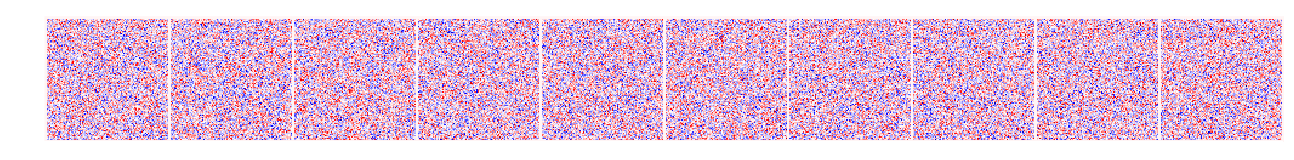}
    \caption{Top ten matched features under the spatially IID prior of \citet{knowles2007infinite} (final MCMC sample). Colour normalisation different to (a).}
    \label{fig:traceissfa}
  \end{subfigure}
  \caption{Comparison of matched features. Matching done by cosine similarity after normalisation.}
  \label{fig:featurecomparison}
\end{figure}

We can compare the reconstruction performance of iSSFA model to the reconstruction of the data obtained from PCA / SVD.  We estimate the model order using Minka's method resulting in a setting of $K^*=33$. Analysis of dataset dimensionality is shown in Figure \ref{fig:dimensionality}. 
It is of note that Minka's method still provides empirically better PCA performance than using the true $K=50$.

We find that, on the holdout data, the reconstruction error using the first $K^*=33$ principal eigenvectors is nearly 1.67 times greater than the MCMC reconstruction error $ \sum_{t,k} (\bold X_{t,k} - \hat{\bold X}^{iSSFA}_{t,k})^2 $. For the PCA is the solution is a matrix factorisation $\bold X^{EIG} = \bold W^{EIG} \bold S^{EIG}$, where the rows of $\bold S^{EIG}$ are the first fifty eigenvectors and the corresponding PCs. 

We also compare the iSSFA results to another standard matrix decomposition method ICA, implemented through the FastICA algorithm (\citet{hyvarinen2000ICAalgos}). This methodology builds on the PCA solution by further factorising the sources $\bold W^{EIG}=\bold W^{ICA}\bold U^{ICA}$. Substituting this into the PCA decomposition of the data $\bold Y = \bold W^{ICA}\bold U^{ICA} \bold S^{EIG} = \bold W^{ICA} \bold S^{ICA}$ (note that we have changed notation from \citet{hyvarinen2000ICAalgos}). A comparison of these different sets of features with the true features is shown in Figure \ref{fig:featurecomparison} where it is visible that our nonparametric method finds features that are more similar to the true features. 

\begin{figure}[htp]
  \begin{subfigure}{0.5\textwidth}
   \includegraphics[width=1.0\textwidth]{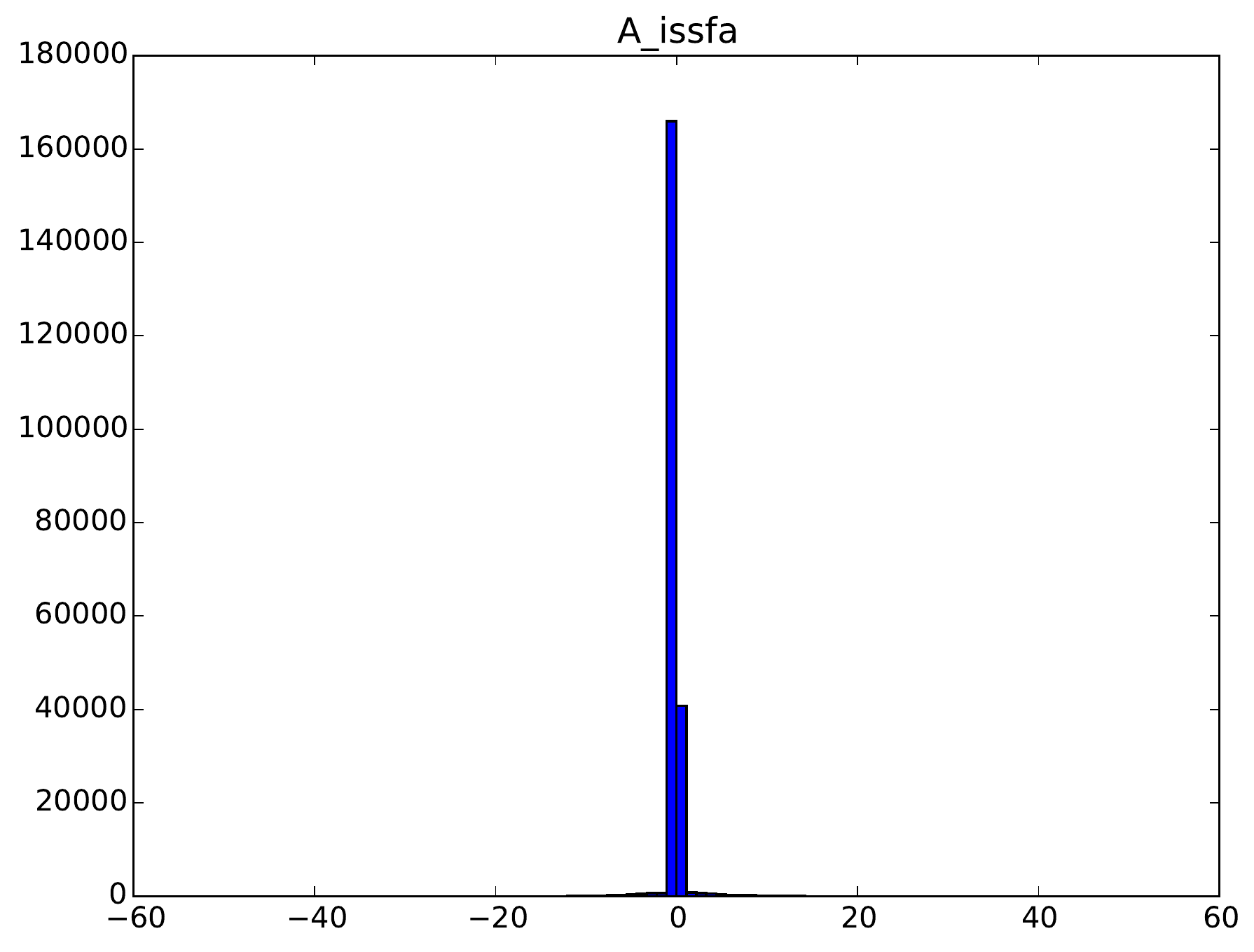}
   \caption{Histogram of sources $\bold W^{iSSFA} = \hat{(\bold A\circ \bold Z)}^{iSSFA}$ in the final MCMC iteration, excluding (two) unique features ($\sum_t \bold Z_{t,k}=1$). Sample excess kurtosis: 73.8. The distribution is highly non-Gaussian owing to the spike-and-slab type model. Here the slab is extremely thin due to the level of sparsity.}
    \label{fig:kurtosis_issfa}
  \end{subfigure}
  \begin{subfigure}{0.5\textwidth}
   \includegraphics[width=1.0\textwidth]{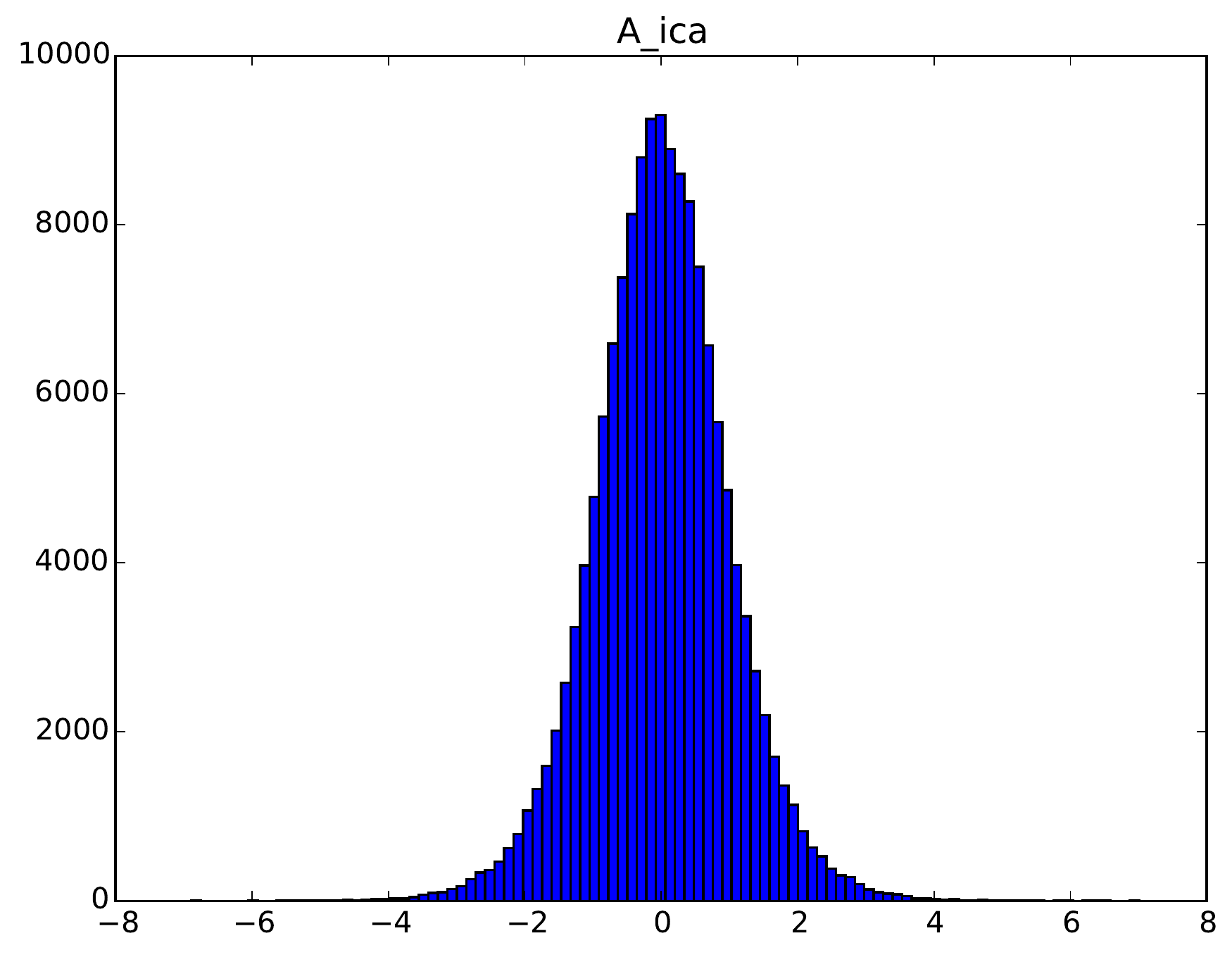}
   \caption{Histogram of sources $\bold W^{ICA}$ from FastICA. Sample excess kurtosis: 1.1}
    \label{fig:kurtosis_ica}
  \end{subfigure}
  \caption{Distribution of inferred sources under iSSFA and FastICA. If one were interested in maximising the non-Gaussianity of inferred sources the sparsity of the iSSFA model produces a more optimal fit by the kurtosis criterion than does the FastICA methodology.}
  \label{fig:kurtosis}
\end{figure}

FastICA attempts to decompose a signal into independent components by maximising the non-Gaussianity of inferred sources $\hat{\bold W}$, using a metric based on an approximation to negentropy. As can be seen in Fig \ref{fig:kurtosis} iSSFA produces more highly non-Gaussian estimates of the source distribution than does the FastICA algorithm applied to the same dataset. This is due to the sparsity induced by the Indian Buffet Process, which turns the iSSFA source distribution into a spike-and-slab model. Using sample excess kurtosis as a measure of non-Gaussianity (higher is more super-Gaussian), in the last MCMC iteration iSSFA scored 73.8 while the corresponding FastICA score was 1.1. The shapes of the histograms indicate that this ranking would hold under other reasonable metrics.

Another metric for the quality of the blind source separation is defined in \citet{knowles2011nonparametric}, essentially $E_r(A,B)=\sum_{k=1}^{K^{TRUE}} \min_{j} \|A_{:,k} - B_{:,j}\|^2  $. Here we found the ratio of errors to be $E_r(\bold S, \bold S^{ICA})$ $/E_r(\bold S, \bold S^{ISSFA})= 1.41$. Whereas under the spatially IID nonparametric prior the same ratio was 0.87, i.e. worse than FastICA. This suggests that, where appropriate, modelling spatial continuity will help to provide a more accurate blind source separation; and also that the number of extra features in the iSSFA model is unlikely to be a sufficient explanation for its better score than ICA.

Figure \ref{fig:features} shows four rows of features. The top row visualises the first true ten latent features. The following three rows show other types of features matched by cosine similarity: iSSFA features from the final MCMC iteration; FastICA features; and finally PCA features. Figure \ref{fig:featurecomparison} shows matched features obtained under the isFA prior of \cite{knowles2007infinite}. We can see that modelling latent covariance structure has allowed the model to find a solution which more closely matches the ground truth than FastICA, which produces noisy features that vary on the wrong spatial scale. We also see that even within nonparametric models, accounting for latent covariance has improved feature recovery relative to the IID prior.

Hence, at least when considering sparsely activated data, nonparametric sparse factor analysis can provide a more accurate reconstruction of the original data than can the eigendecomposition. Furthermore relaxation of orthogonality constraints can lead to features which are, individually, more similar to latent features.

\section{Neuroimaging application}
\label{sec:neuroexample}

In this example we consider application of the model to fMRI neuroimaging data, demonstrating the abiity to produce sensible results at scale. The dataset contains 30 subjects from the Cambridge Centre for Ageing and Neuroscience study (\citet{campbell2015idiosyncratic,shafto2014cambridge}). Subjects watched an editted 8 minute video of a Hitchcock drama \textit{Bang! You're Dead}. For each subject this yielded 193 images for a total of $T=193 \times 30=5790$, which after preprocessing had a spatial resolution of $V=61 \times 73 \times 61 = 271633$. 

For fMRI data, naive modelling of spatial dependencies using dense matrices would be infeasible due to the need for inversion and determinants of $V \times V$ covariance matrices. Hence efficient computation is necessary if we wish to model spatial structure. Spatial modelling is well motivated by the fact that the spatial resolution of fMRI is finer than typical parcellations of the brain into functional regions, and hence we should expect neighbouring voxels to often share similar values. Furthermore, as this type of task involves viewing `naturalistic' video, as opposed to an experimental block design, analysis methods must be relatively unconstrained in terms of the form of the time courses and brain regions involved. Hence the model presented is well suited to the requirements of the analysis task.

\def \run_uid {2016-11-23T17-09-01.313}
\def \noteone {overlay}
\def \typeone {axial}
\def \notetwo {slice}
\def \typetwo {sagittal}

\begin{figure}
  \begin{subfigure}{0.5\textwidth}
   \includegraphics[width=1.0\textwidth]{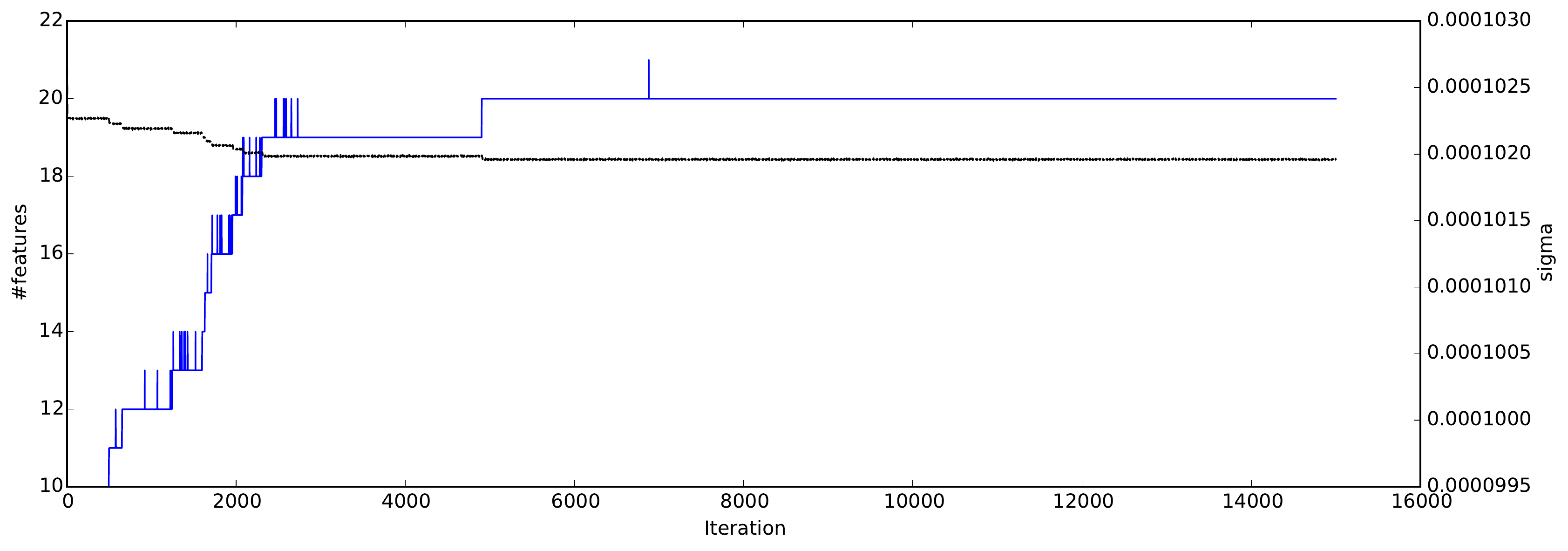}
    \label{fig:khistory}
  \end{subfigure}
  \caption{Trace plot of the number of features and noise level over MCMC iterations.}
  \label{fig:neurotrace}
\end{figure}

We ran the sampler using 6 compute nodes with for 15000 iterations. We used a prior distribution on the noise level which places a soft constraint on the desired reconstruction accuracy. Trace plots of the number of features and inferred noise level are shown in Fig \ref{fig:neurotrace}, where the chain appears to converge.

\begin{figure}
  \begin{subfigure}{0.47\textwidth}
   \includegraphics[width=1.0\textwidth]{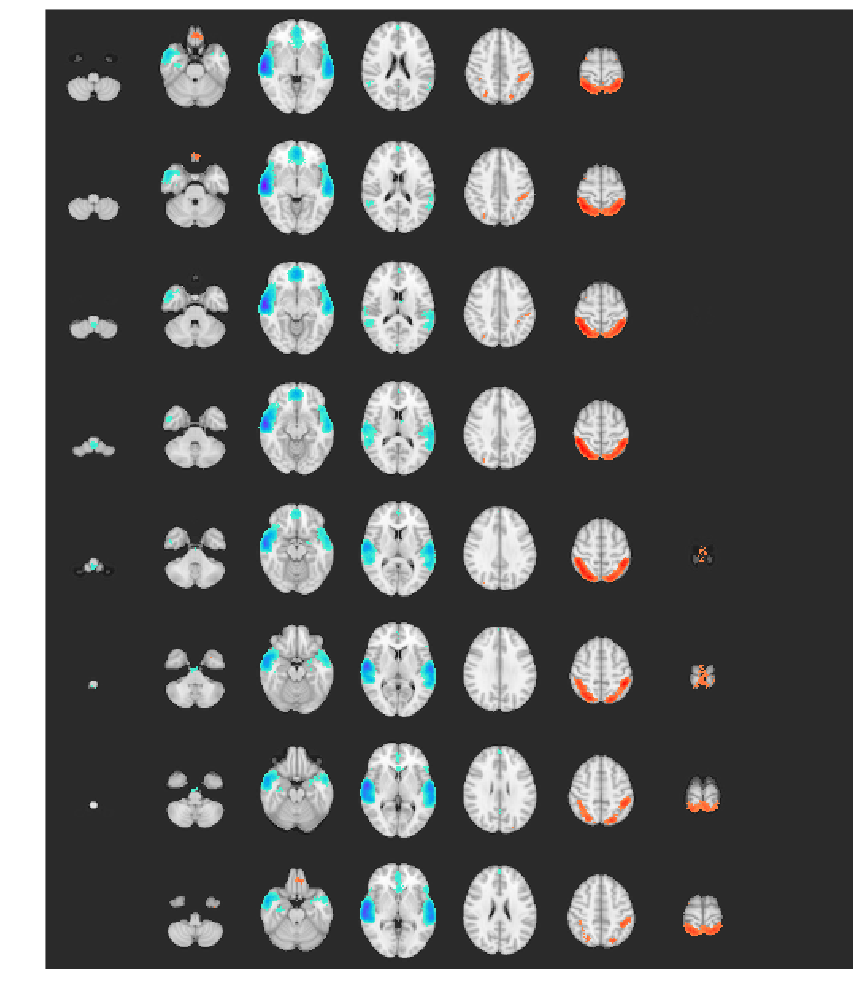}
    \label{fig:component_axial_3}
  \end{subfigure}
  \begin{subfigure}{0.47\textwidth}
    \includegraphics[width=1.0\textwidth]{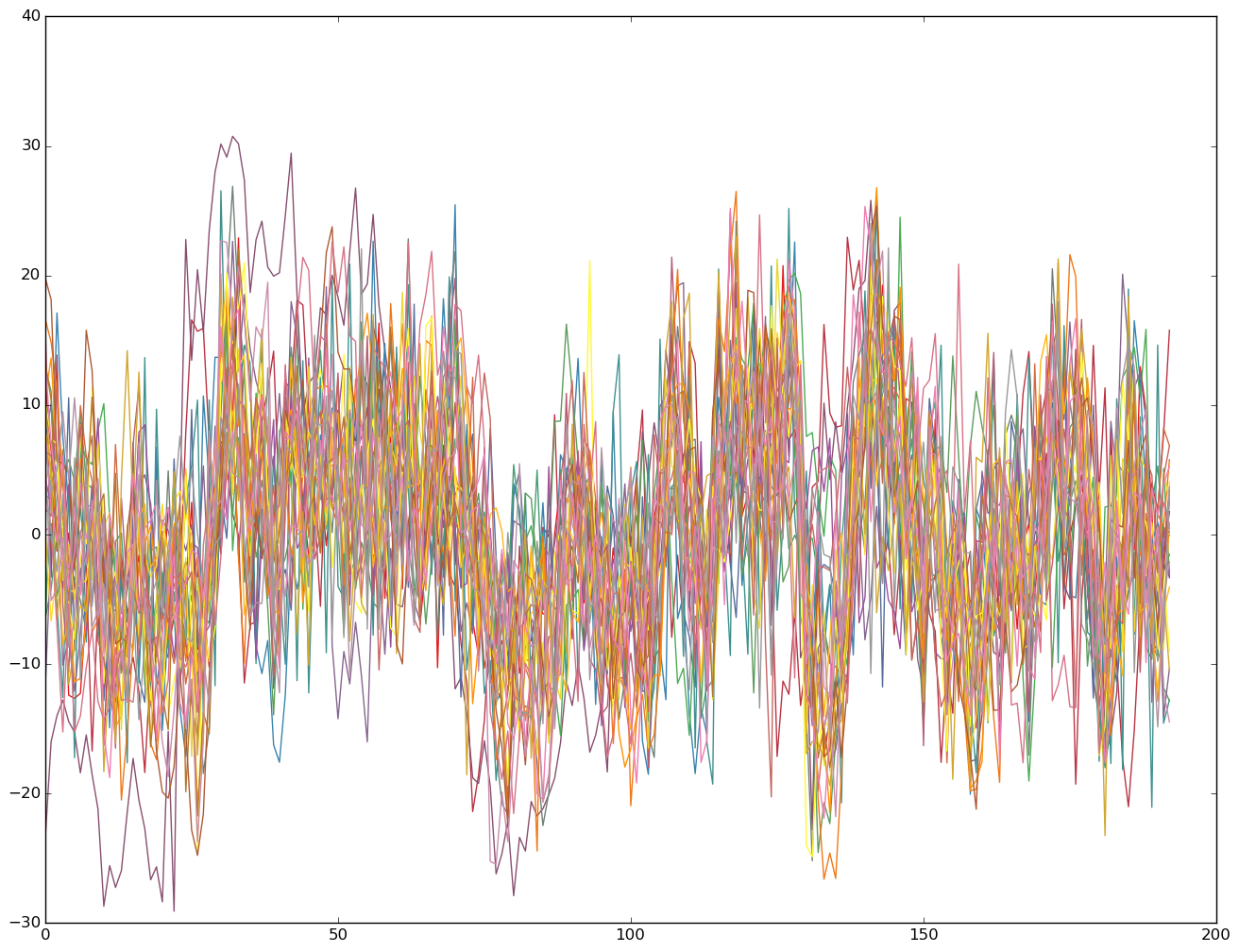}
    \label{fig:subject_timeseries_3}
  \end{subfigure}   
  \caption{The component with highest inter-subject weight correlation. Heavy loadings are on the auditory region, and with opposite sign on the parietal lobe.}
  \label{fig:comp3}
\end{figure} 

\begin{figure}
  \begin{subfigure}{0.47\textwidth}
   \includegraphics[width=1.0\textwidth]{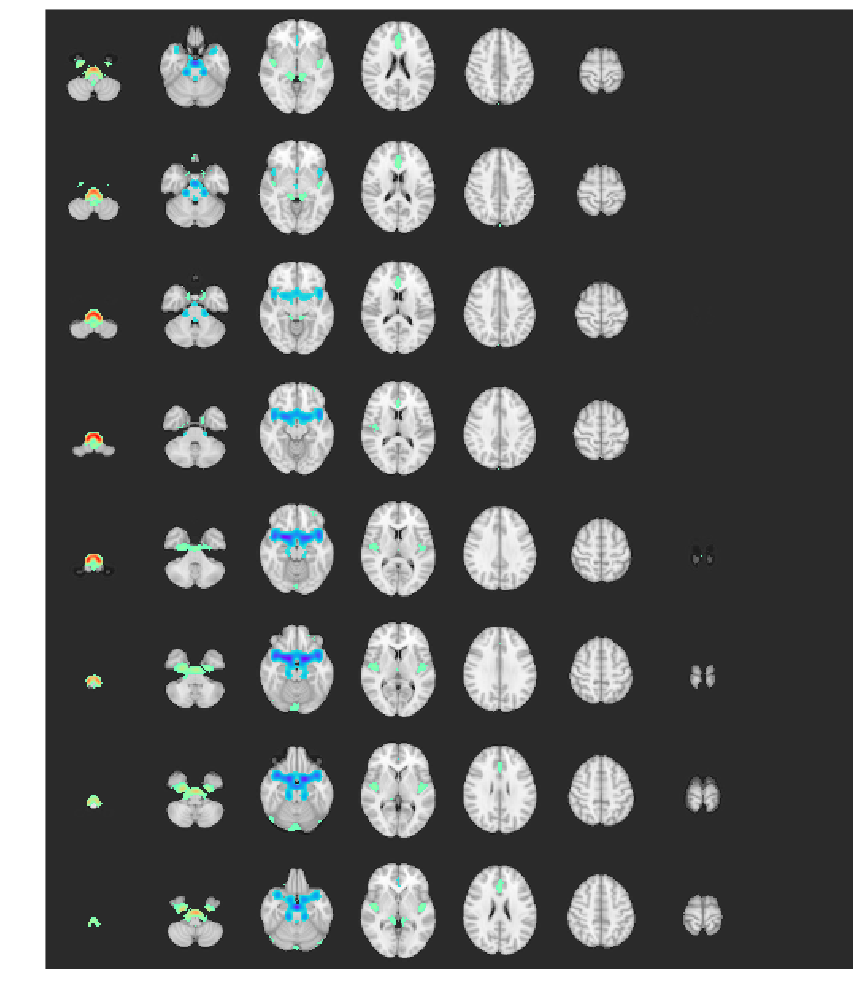}
    \label{fig:component_axial8}
  \end{subfigure}
  \begin{subfigure}{0.47\textwidth}
    \includegraphics[width=1.0\textwidth]{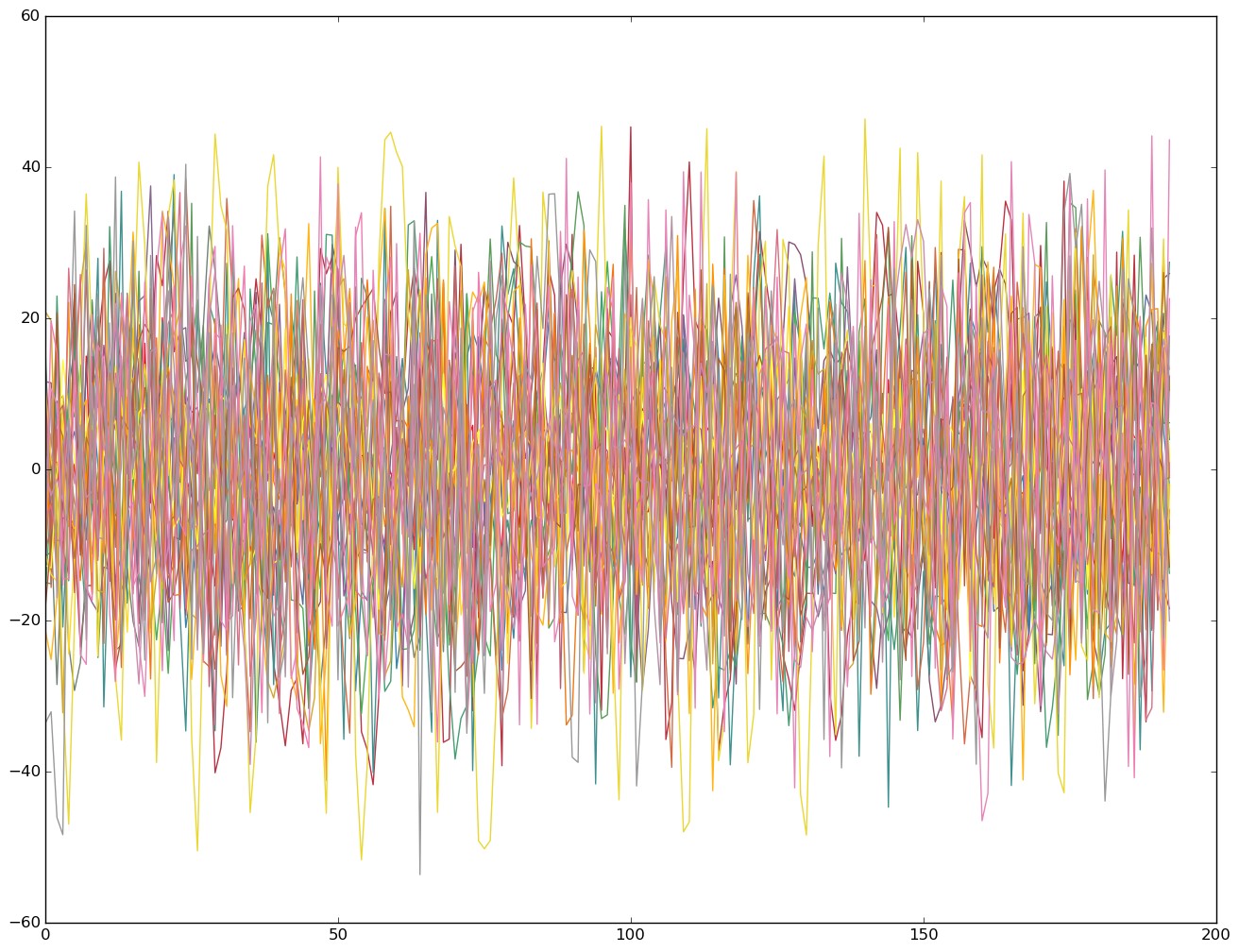}
    \label{fig:subject_timeseries_8}
  \end{subfigure}   
  \caption{The component with lowest inter-subject weight correlation, where the accompanying spatial vector is weighted heavily toward the brain stem.}
  \label{fig:comp8}
\end{figure} 

Two selected components from the final MCMC sample are visualised in Figure \ref{fig:comp3} and Figure \ref{fig:comp8}. The latent spatial vectors are thresholded at $(2/V)\sum_v |\bold S_{k,v}|$ and overlaid on an anatomical reference image. The time courses for each subject are also visualised by slicing the $\bold A_{:,k}$ appropriately.

We can measure the correlation between different subject-wise time-series as a measure of synchronisation, which is often seen as indicative of a link to the experimental task. The component shown in \ref{fig:comp3} has a very high intra-subject correlation. Visual inspection of the spatial vector shows the component loads heavily on auditory regions as well as the parietal cortex. The accompanying subject-wise time series show a strong level of synchronisation. Hence this component is biologically plausible.

The feature in Figure \ref{fig:comp8} appears more like an artefact in the data and had the least synchronised set of time courses in the decomposition. We can see that the component primarily loads very heavily on the brain stem, and so is unlikely to relate meaningfully to the experimental task.

\section{Discussion}
\label{sec:discussion}


We have extended methods for matrix factorisation with an unknown number of latent vectors. We imposed soft constraints on the regularity of latent features through a nondiagonal latent covariance structure. We provided a roadmap for combinining different forms of regularity constraints. We implemented in detail a model for smoothness of latent features expressed through an eigendecomposition based on the Discrete Cosine Transform. The method is general however, and for example, we could penalise activity in particular regions using the Discrete Wavelet Transform. 

The model presented here also improves on previous work by allowing non-zero means for feature weights, which enables the learned feature vectors to explain variation seen in the data, rather than relying on covariance structure.  

Section \ref{sec:simulation} demonstrated the benefits of this modelling strategy in a denoising example when the latent features were smooth 2D images. It was seen that this improved the mixing of the chain, recovered features closer to the ground truth, and removed noise more effectively than the prior used in previous work. As the latent activations were sparse we also compared our inferred weights to the output of FastICA, observing that the model presented here learned a more highly kurtotic weight distribution thus optimising this ICA proxy criterion more effectively.

In Section \ref{sec:neuroexample} we demonstrated these methods on a neuroimaging application. The efficient computational strategies introduced here for dealing with nondiagonal latent covariance were essential since numerical methods would fail on $V \times V$ matrices when $V>10^5$. We are not aware of other methods that are able to provide similar modelling capabilities at present.


\section{Ackowledgements}

This work was supported by the Medical Research Council [Unit Programme number U105292687]. The Cambridge Centre for Ageing and Neuroscience (Cam-CAN) research was supported by
the Biotechnology and Biological Sciences Research Council (grant number BB/H008217/1).


\label{sec:bibliography}
\bibliographystyle{plainnat}
\bibliography{bib_ica,bib_deep,bib_neuro}

\end{document}